%% file: main.tex
\documentclass[10pt,twocolumn,letterpaper]{article}
\usepackage[pagenumbers]{preprint}
\usepackage{enumitem}
\usepackage{algorithm}
\usepackage[noend]{algpseudocode}
\usepackage{array,amsmath,amssymb,amsthm,booktabs,multirow,graphicx,xcolor,float,adjustbox}
\newcommand{\fitcol}[1]{\begin{adjustbox}{max width=\columnwidth}#1\end{adjustbox}}
\newcommand{\fittext}[1]{\begin{adjustbox}{max width=\textwidth}#1\end{adjustbox}}
\definecolor{linkblue}{rgb}{0.21,0.49,0.74}
\usepackage[breaklinks,colorlinks,allcolors=linkblue]{hyperref}
\hypersetup{pdftitle={Driving World Models Move the Camera, Not the Scene},pdfauthor={Yuqiao Feng, Xiaoyu Ao}}
\title{Driving World Models Move the Camera, Not the Scene}
\author{Yuqiao Feng\quad Xiaoyu Ao\qquad Chongqing University}
\begin{document}
\twocolumn[{
  \renewcommand\twocolumn[1][]{#1}
  \maketitle
\begin{minipage}{\textwidth}
\centering\includegraphics[width=\textwidth]{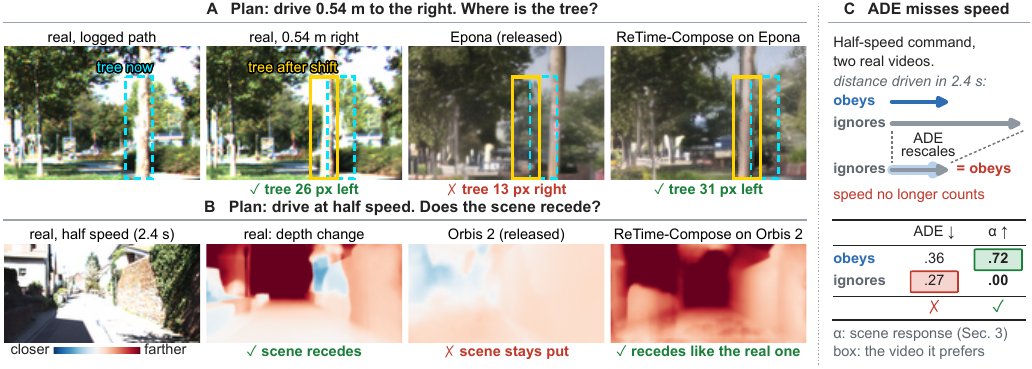}
\captionof{figure}{\textbf{Released driving world models move the camera, not the scene, and the trajectory score cannot tell.} A: the stereo right camera records the real outcome of a plan 0.54\,m to the right (cyan: the tree under the logged path; yellow: where it must move). Released Epona moves it the wrong way; ReTime-Compose on the same frozen Epona follows (unselected KITTI context; all sets in Table~\ref{tab:lateral}). B: at half speed the real scene recedes (depth change against the logged speed, from the retimed log); released Orbis~2's barely changes, ReTime-Compose's recedes (selected for visibility; all contexts in Sec.~\ref{sec:exp}). C: two real videos under a half-speed command, one that obeys it and one that ignores it (arrows: distance driven, schematic). The scale-aligned ADE of current benchmarks rescales each estimated path before comparing, which removes the speed difference, and favours the video that ignores the command; the scene response $\alpha$ (Sec.~\ref{sec:refs}) does not (92 contexts, Fig.~\ref{fig:dose}).}
\label{fig:reference}
\end{minipage}
\vspace{8pt}
}]
\input{sec0_abstract}
\input{sec1_intro}
\input{sec2_related}
\input{sec3_analysis}
\input{sec4_retime}
\input{sec5_experiments}
\input{sec6_limits}
\section*{LLM Usage Statement}
We used Claude (Claude Code) and GPT models (ChatGPT, Codex) to assist with drafting, polishing and experiment code. The authors determined the research questions, analyses, decision criteria and conclusions, revised the text, checked every number against the logged outputs and take full responsibility.

{\small\bibliographystyle{ieeenat_fullname}\bibliography{refs}}
\clearpage
\appendix
\begin{center}{\Large\bf Supplementary Material}\end{center}
\noindent\textbf{How this supplement is organized.} Appendices~\ref{app:repair}--\ref{app:compose} detail the context repair and ReTime-Compose (Secs.~\ref{sec:channel}--\ref{sec:exp}), including the speed response without a depth reader; Appendices~\ref{app:metric} and~\ref{app:noise} detail the diagnosis of Sec.~\ref{sec:diagnosis}. Appendix~\ref{app:retimeresults} covers fine-tuning on retimed pairs; Appendices~\ref{sec:method} and~\ref{app:gap} the small geometric forecaster of Sec.~\ref{sec:predictor}, which is not part of ReTime-Compose; Appendix~\ref{app:chain} links each design choice of the context repair to its ablation.
\input{supp_compose}
\input{supp_diag}

\input{supp_forecaster}
\end{document}

%% file: sec0_abstract.tex
\begin{abstract}
Plan-conditioned driving world models are used to compare candidate plans, so their forecast should change with the plan as the real scene would, yet widely used control metrics check only whether the camera follows the command. Recorded drives already contain real outcomes of plans that were never driven: retiming a log gives pose-matched images and LiDAR for other speeds, and a stereo rig records a laterally shifted view. Against these references, released world models move their camera but barely move the scene: their scene response, the share of the real change in distance that the forecast reproduces, reaches at most about a third of real video's, and the trajectory score cannot tell an executed speed command from an ignored one. Interventions inside the generator trace the failure to the plan's channel. The forecast geometry follows the context frames; once they carry the planned geometry, plan tokens add little. ReTime-Compose therefore delivers the plan as geometry, without training. It transports the frames of one logged-speed rollout to the planned cameras, and the frozen model completes only what no frame covers. Frozen generators then follow a 0.54\,m lateral shift nearly as the real scene does on KITTI, nuScenes and Argoverse~2, and respond more than re-sampling to slow-down plans in three generators and to speed-up plans in two.
\end{abstract}

%% file: sec1_intro.tex
\section{Introduction}
\label{sec:intro}

A planner that asks a driving world model ``what if I slow down here?'' expects a forecast in which the parked car ahead stays farther away than under the logged speed. Driving world models are increasingly used this way: given the past frames and a plan, the future ego trajectory, they forecast what the vehicle will see, and candidate plans are compared on these forecasts~\cite{gem,vista,epona,orbis2}. For this use the forecast must change with the plan, as the real scene would: a forecast whose camera slows down while its scene does not would tell the planner that slowing down changes nothing.

Widely used control metrics check this through the camera: ACT-Bench~\cite{actbench}, DrivingGen~\cite{drivinggen} and Orbis~\cite{orbis,orbis2} recover the ego path from the generated video and compare it with the command (Table~\ref{tab:evals}), and DrivingGen itself notes that no metric checks the scene under the commanded path. The scene goes unchecked because the outcome of a plan that was not driven is not recorded. So it is open: \emph{does the forecast scene move with the plan as the real one would?}

We observe that two families of such outcomes are in fact recorded (Fig.~\ref{fig:reference}). Along a recorded path, arriving earlier or later at a pose is itself recorded, so retiming the log supplies pose-matched images and LiDAR for alternative speed plans. A stereo rig records, at the same instant, the view of a camera shifted laterally by its baseline. Against these references, released world models move their camera but not their scene. Given a half-speed plan, Epona's generated camera covers the planned path (1.18 times its length), yet the distance to the scene ahead changes about a quarter as much as in real video, as if the scene travelled along with the camera. We quantify this with the \emph{scene response} $\alpha$, the share of the real change in distance to static scene points that the forecast reproduces, read against LiDAR (Fig.~\ref{fig:alpha}). Released models reach at most about a third of the real-video response under speed plans and at most a quarter of the real image motion under a 0.54\,m lateral shift (Table~\ref{tab:twolayer}). The trajectory score these metrics report (scale-aligned average displacement error, ADE) cannot see this failure, because its scale alignment absorbs the change of speed: it even prefers a real video that ignores a speed command to one that executes it (Figs.~\ref{fig:reference}C,~\ref{fig:dose}). Across five released models and ten fine-tuned variants, the score does not rank the response.

Why does the scene not follow? The plan enters the model as tokens, an embedding of the commanded motion, but the forecast geometry is set by the context frames, the past frames the model conditions on: a \emph{plan--geometry channel mismatch}, which controlled interventions inside the generator locate one computation at a time. The tokenizer and the sampler keep the real response, and once the context frames carry the plan the model follows it, while replacing the plan tokens by the logged speed then changes little, in DrivingWorld and in Epona. It also explains why two repairs fall short, since repairing the context step by step compounds local errors along the warp chain and fine-tuning on retimed pairs recovers at most half of the real response.

If this diagnosis is right, a frozen model should follow a plan that reaches it as geometry. Moving frames to new cameras and letting a generator complete the rest is an established structure~\cite{cfwm2026,worldforge,gen3c}; what a planner lacks is a source, since it has only the history, while retrospective transport uses the real future. A speed plan retimes one path, so a model's own rollout at the logged speed already passes through the poses of slower and faster plans. ReTime-Compose retimes this rollout as we retime real logs for the reference, transporting the rollout frames to the planned cameras with the depth and poses read from them, and the frozen generator completes only what no frame covers (Fig.~\ref{fig:compose}). Transport places the scene where the plan puts it; the generator supplies what transport cannot, and at a one-lane shift completion fills two fifths of each frame and raises the response from .71 to .76. With one recipe and no per-model tuning, frozen Epona, GEM and Orbis~2 follow lateral plans that the released models cannot follow, and on speed plans outperform \emph{re-sampling}, the training-free baseline that replays the logged-speed rollout at the planned speed, by .11 to .24 in response.

Our contributions are:
\begin{itemize}[leftmargin=*,itemsep=1pt,topsep=2pt]
\item \textbf{Real counterfactual references and the scene response.} Retimed logs and the stereo pair give real outcomes of undriven speed and lateral plans; the scene response $\alpha$ and its noise read a forecast against them, which shows that the standard trajectory score cannot tell an executed command from an ignored one (Secs.~\ref{sec:refs},~\ref{sec:metric}).
\item \textbf{A mechanistic diagnosis.} Controlled interventions in two generators localize the failure to a plan--geometry channel mismatch (Sec.~\ref{sec:channel}).
\item \textbf{A diagnosis-guided repair.} ReTime-Compose delivers the plan as geometry from evidence available at decision time, without training, with a pose-only criterion for where it pays off, on KITTI~\cite{kitti,kittidepth}, nuScenes~\cite{nuscenes}, Argoverse~2~\cite{av2} and three generators (Secs.~\ref{sec:compose},~\ref{sec:exp}).
\end{itemize}

%% file: sec2_related.tex
\section{Related work}
\label{sec:related}
\begin{figure*}[!t]
\centering\includegraphics[width=\textwidth]{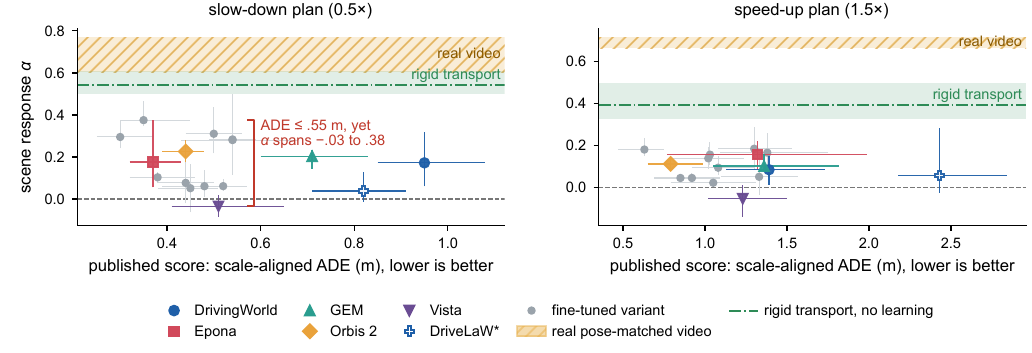}
\caption{\textbf{The trajectory score does not rank the response to a plan.} KITTI. Coloured: released world models (hollow: DriveLaW, exploratory adapter); grey: fine-tuned variants. Horizontal: Orbis~2's scale-aligned ADE~\cite{orbis2}; vertical: scene response $\alpha$ (Eq.~\ref{eq:direct-response}). Hatched yellow band: real pose-matched video; dash-dot line: rigid transport of the context depth, no learning. Red bracket: models with nearly the same score span the whole range of response. Bars: 95\% drive-cluster intervals.}
\label{fig:metric}
\end{figure*}
\paragraph{Why current evaluations miss the failure.} Current evaluations read the ego path from generated video or score novel views without ground truth, so a camera that follows the command passes them (Table~\ref{tab:evals}). ACT-Bench regresses the ego action with a trained estimator~\cite{actbench}, DrivingGen and VehDyn recover trajectories by SLAM or visual odometry~\cite{drivinggen,vehdyn}, and Orbis~2 and HelloWorld fit a scale to the estimated path before the error~\cite{orbis2,helloworld}, an error also used as a reward~\cite{georeward}. WorldLens and What-If World score novel trajectories or interventions without ground truth~\cite{worldlens,whatifworld}; EponaV2 decodes forecast depth but evaluates planning only~\cite{eponav2}. PlayWorld argues that trajectory scores can confound geometry and control failures~\cite{playworld}; Su \etal report that a driving model's video does not follow a trajectory steered at sampling time~\cite{norminject}, the two-layer failure we measure. None compares generated geometry with real sensors under an unrecorded plan; Sec.~\ref{sec:metric} does.

\paragraph{Why existing references do not settle it.} Exact references for plans that were not executed come from simulation or from real frames of the target, which a planner does not have. Zhang \etal obtain exact counterfactual references in CARLA, transporting the factual continuation into the counterfactual view and filling the rest with a frozen model~\cite{cfwm2026}; BranchDrive executes alternative actions in CARLA~\cite{branchdrive}; World in World lets a frozen generator regenerate a target view from geometry rendered from real source video~\cite{worldinworld}. ReTime-Compose shares this structure but takes its sources from the model's own logged-speed rollout, at a measured cost against the real future. Novel-trajectory synthesis for driving~\cite{freevs,streetcrafter,drivedreamer4d,recondreamer,geov2v} renders shifted lanes from the whole log without ground truth; we forecast from the history and check against the right camera. Para-Lane registers parallel real traversals for lateral novel views~\cite{paralane}, and KITScenes scores lateral shifts against projected map signs~\cite{kitscenes}; retiming, a temporal resampling that also supervises speed in generic video~\cite{fastslow}, is their longitudinal, single-log counterpart.

\paragraph{Why explicit geometry alone does not solve it.} Explicit geometry is known to help camera control, but at decision time it must come from the history alone. GEN3C conditions generation on a rendered 3D cache because ``camera parameters are mere inputs''~\cite{gen3c}; GeoDrive renders a first-frame point cloud along the ego trajectory and scores control by re-estimated ADE~\cite{geodrive}; FWD warps depth features~\cite{fwd}; WorldForge and NVS-Solver guide a frozen model with warped views~\cite{worldforge,nvssolver}; Warp-as-History feeds warped past frames through the history pathway~\cite{warpashistory}, the training-free form of our context repair (Sec.~\ref{sec:channel}); WorldWarp and CamTrol++ bound the error of long or large camera motion~\cite{worldwarp,camtrolpp}. From the history alone, transporting one frame along the whole plan is noisy (App.~\ref{app:compose}) and sampling on one's own output compounds errors~\cite{gamengen,selfforcing,svi}, so ReTime-Compose never re-warps a warped frame. Robot latent actions absorb visual context~\cite{cdlam}; we add, inside a released token model and against real LiDAR, that the plan token carries no geometric effect once the context is right.

%% file: sec3_analysis.tex
\section{Released models move the camera, not the scene}
\label{sec:diagnosis}
\begin{figure*}[!t]
\centering\includegraphics[width=\textwidth]{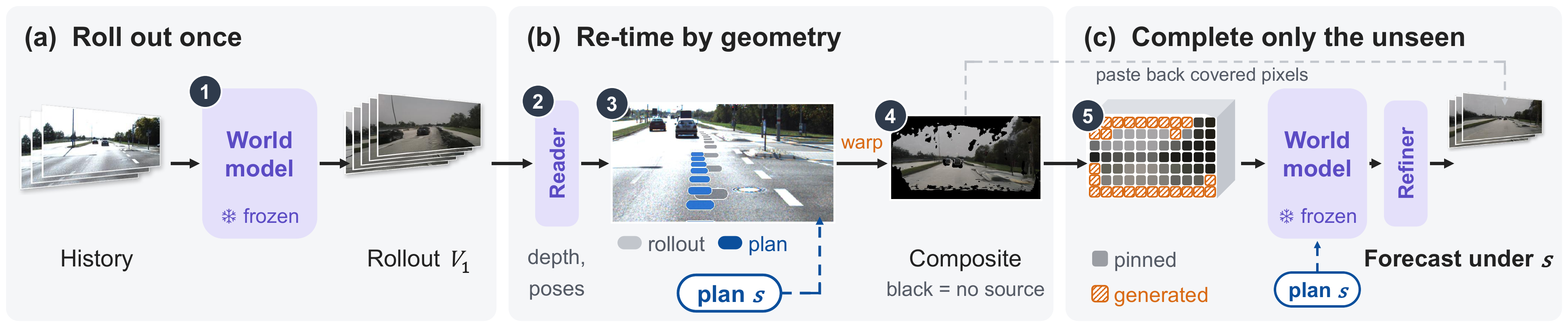}
\caption{\textbf{ReTime-Compose: the plan acts through transport, and the frozen generator completes only what transport cannot supply.} (a) One frozen rollout at the logged speed. (b) Read depth and poses transport its frames from the rollout cameras (gray) to the planned ones (blue); the plan's geometry enters only here. (c) Covered tokens are pinned, the rest generated; a refiner reduces artefacts.}
\label{fig:compose}
\end{figure*}
A plan $p$ specifies future camera poses; a video world model generates frames under $p$ from a history $H$ of past images and poses, and we read their geometry with a frozen metric depth reader. Logs alone cannot tell whether a model uses the plan: if the logged plan is $p=\pi(H)$, the plan-blind predictor $\widetilde F(H,p)=F(H,\pi(H))$ has the same logged risk as $F$ (on KITTI, past speed predicts the next 2.4\,s speed with $R^2{=}.90$). A reference for \emph{undriven} plans is needed.

\subsection{Real references for other plans}
\label{sec:refs}
\paragraph{Speed.} Let $T(t)$ be the recorded camera trajectory and $\tau$ monotone. Retiming defines the plan $p_j=T(\tau(j))$ with the history unchanged; for static points the recorded image and LiDAR at $\tau(j)$ are the reference at that pose (Fig.~\ref{fig:qual05}). For example, at half speed ($\tau(j)=j/2$) the planned camera at step 24 is where the car really was at step 12, so the frame and LiDAR recorded then show what the half-speed plan would see. It holds for static content; moving points are dropped on nuScenes and, on KITTI, enter all plans alike.
\paragraph{Lateral.} A rectified stereo rig records, at the same instant, a camera displaced by the baseline $B$ (0.54\,m on KITTI). A plan that shifts right by $B$ asks for exactly the right camera, a reference that also holds for moving objects; stereo pairs are a standard target for single-image view synthesis~\cite{tucker2020}. A pure shift moves a point by the disparity $-fB/Z$, so we regress the horizontal optical flow between the generations of the logged and the shifted plan on $-fB/Z$ from LiDAR; no learned reader is involved on the truth side.
\begin{figure}[t]
\centering\includegraphics[width=\columnwidth]{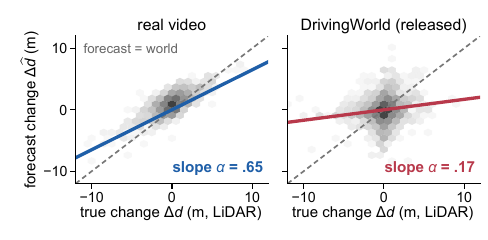}
\caption{\textbf{How the scene response $\alpha$ is read.} Half against logged speed, KITTI; each point is a near-field LiDAR point at one step (6000-point sample). $\alpha$ is the least-squares slope (lines: all points of 59 contexts); dashed: a forecast that changes as the world does. Real video reads .65, not 1, as the depth reader compresses depth.}
\label{fig:alpha}
\end{figure}
\paragraph{Scene response and noise.} We ask how much of the real change of the scene a forecast reproduces when the plan changes. Take two plans $p$ and $q$, e.g.\ half and logged speed, at the same forecast step. On near-field points (both LiDAR depths within 25\,m) let $\Delta d=d_p-d_q$ be the true depth change between the plans and $\Delta\widehat d$ the change between the two forecasts. A forecast that follows the plan as the world does has $\Delta\widehat d=\Delta d$; one that ignores the plan has $\Delta\widehat d\approx0$. The response is the slope of $\Delta\widehat d$ on $\Delta d$ (Fig.~\ref{fig:alpha}) and the noise the scatter around it:
\begin{align}
 S_{p,q}&=\textstyle\sum(\Delta\widehat d-\Delta d)^2, \label{eq:paired}\\
 \alpha&=\frac{\sum\Delta\widehat d\,\Delta d}{\sum(\Delta d)^2},\quad
 N=\frac{S_{p,q}}{\sum(\Delta d)^2}-(1-\alpha)^2 .
 \label{eq:direct-response}
\end{align}
We call $\alpha$ the \emph{scene response}: the share of the true change in distance that the forecast reproduces, 0 if the plan is ignored and 1 if the forecast changes as much as the world. Unlike a trajectory score, it reads the scene, not the camera, and against a recorded outcome, not the command. The reader compresses depth (within a frame its depth against LiDAR has slope .77), so real video reads about .72, the attainable reference reported next to every model. We call $N$ the \emph{noise}, the variance of the forecast change orthogonal to the true change, and $S_{p,q}/\sum(\Delta d)^2=N+(1-\alpha)^2$ the \emph{total error} ($1{-}\mathrm{EF}$ in the appendix; EF: explained fraction). Intervals are 95\% drive-cluster bootstraps, and every decision rule was fixed before the reading it decides.

\subsection{The trajectory score misses the scene}
\label{sec:metric}
\begin{table}[t]
\centering\footnotesize\setlength{\tabcolsep}{3pt}
\caption{\textbf{What control evaluations compare.} Without the real outcome of an undriven plan, they check the camera, not the scene.}
\label{tab:evals}
\begin{tabular}{@{}lll@{}}
\toprule
Evaluation & compares & real outcome\\
\midrule
ACT-Bench~\cite{actbench} & ego action vs.\ command & none\\
DrivingGen~\cite{drivinggen}, Orbis~2~\cite{orbis2} & ego path vs.\ command & none\\
CameraCtrl~\cite{cameractrl}, AC3D~\cite{ac3d} & camera vs.\ command & none\\
WorldLens~\cite{worldlens} & planner, geometry & logged plan\\
\midrule
$\alpha$ (ours) & scene vs.\ LiDAR & retimed, stereo\\
\bottomrule
\end{tabular}
\end{table}
\begin{figure}[t]
\centering\includegraphics[width=\columnwidth]{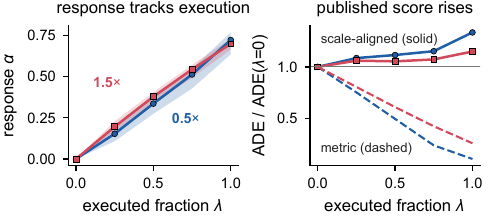}
\caption{\textbf{The scale-aligned score does not fall as a real video executes more of the command; the response does.} 92 KITTI contexts. Left: response $\alpha$ against the executed fraction $\lambda$. Right: Orbis~2's scale-aligned ADE (solid) and the unaligned error (dashed); App.~\ref{app:metric}.}
\label{fig:dose}
\end{figure}
\paragraph{The score is blind to execution.} The trajectory score cannot tell whether a speed command was executed, because its scale alignment removes exactly the quantity it should measure. We build real videos that execute a known fraction $\lambda$ of a speed command by taking logged frame $\mathrm{round}((1+\lambda(s-1))j)$ at step $j$. As a monotonicity test~\cite{rethinkfid}, the response grows linearly with $\lambda$, whereas Orbis~2's scale-aligned ADE~\cite{orbis2} does not fall as more of the command is executed (Fig.~\ref{fig:dose}). Without the least-squares scale the error falls from 4.6 to 0.5\,m.

\paragraph{Fifteen models.} Across five released models (DrivingWorld~\cite{drivingworld}, Epona, GEM, Orbis~2, Vista) and ten variants fine-tuned on logged, speed-augmented or retimed pairs, the score does not rank the response (Fig.~\ref{fig:metric}; Kendall rank correlation $+.16$ and $-.03$). Fine-tuning on logged pairs improves the score of DrivingWorld and Orbis~2 while their slow-down response falls, and in 13 of 16 KITTI models (DriveLaW~\cite{drivelaw} included) and all six nuScenes models a model's own logged-speed video scores \emph{better} against the slow-down command than its slow-down video. Fine-tuning on retimed pairs helps, but reaches at most half the real-video response (App.~\ref{app:retimeresults}).

\begin{table}[t]
\centering\small\setlength{\tabcolsep}{3.5pt}
\caption{\textbf{Camera execution does not imply scene response.} KITTI, 59--256 contexts per model. $k_s$: length of the generated camera path over the real path of the same plan (1: executed). Scene response $\alpha$ (Eq.~\ref{eq:direct-response}); lateral: slope on real stereo disparity for a 0.54\,m right shift, 32 contexts, two seeds (Orbis~2, GEM: one).}
\label{tab:twolayer}
\fitcol{\begin{tabular}{lcccccc}
\toprule
 & \multicolumn{3}{c}{camera path $k_s$} & \multicolumn{3}{c}{scene response $\alpha$}\\
Model & $0.5\times$ & $1\times$ & $1.5\times$ & $0.5\times$ & $1.5\times$ & lateral\\
\midrule
DrivingWorld & 1.02 & .69 & .63 & .17 & .08 & $-$.30 / $-$.15\\
Vista & .53 & .69 & .52 & $-$.03 & $-$.05 & .03 / .02\\
Epona & 1.18 & 1.17 & 1.15 & .18 & .16 & $-$.03 / $-$.02\\
Orbis~2 & 1.42 & 1.10 & .89 & .23 & .11 & .23\\
GEM & .77 & .83 & .74 & .20 & .10 & $-$.02\\
\midrule
real video & 1 & 1 & 1 & .64--.73 & .66--.71 & .99\\
\bottomrule
\end{tabular}}
\end{table}
\paragraph{The scene lags the camera.} Executing the plan with the camera, the layer a trajectory score reads, does not imply a response of the scene (Table~\ref{tab:twolayer}). In their cameras the models differ, from a speed ceiling (DrivingWorld, Vista) to a constant overshoot (Epona); in their scenes they agree: none reproduces more than .23 of the true change, against .64--.73 for real video. Under the lateral plan, given to each model through its own trajectory interface as the logged path shifted 0.54\,m to the right (Epona's pose actions carry the full displacement), no model exceeds .23, against .99 for real stereo. The forecast distance to the nearest obstacle in the ego corridor fails the same way (App.~\ref{app:metric}). Rigidly transporting the context frame's reader depth along the plan, with no learning, responds with .54 and .39 on the same contexts, two to four times the released models (Fig.~\ref{fig:metric}; a depth map, not a video: Sec.~\ref{sec:cmp}).

\subsection{Where the plan enters}
\label{sec:channel}
\begin{table}[t]
\centering\small\setlength{\tabcolsep}{3pt}
\caption{\textbf{Given the plan's real frames as context, the plan tokens no longer matter.} DrivingWorld, 32 contexts; response $\alpha$ at $0.5\times$/$1.5\times$ (real video .72/.72); one computation changed per row.}
\label{tab:loc}
\fitcol{\begin{tabular}{llc}
\toprule
Step & Intervention & $\alpha$\\
\midrule
--- & unmodified, free & .19 / .09\\
tokenizer & VQ round trip of real future & .65 / .64\\
sampling & greedy / $T{=}0.5$ & .21 / .05; .08 / .09\\
context & real frames of the plan as context & .61 / .56\\
plan tokens & same, tokens at logged speed & .61 / .59\\
\bottomrule
\end{tabular}}
\end{table}
The plan enters the model as tokens, but the forecast geometry is set by the context frames.

\paragraph{The tokenizer and the sampler keep the response.} Table~\ref{tab:loc} changes one computation at a time inside DrivingWorld: real futures passed through its tokenizer keep the real response, and neither greedy nor tempered sampling raises the model's own.

\paragraph{The plan tokens leave the geometry unchanged.} Given the plan's real frames as context at every step, DrivingWorld responds, and replacing the plan tokens by the logged speed then changes little (Table~\ref{tab:loc}). Epona behaves the same way: given the plan's real frames it responds (.63/.60), and logged-speed tokens change this by at most .01, intervals including zero (55 contexts). In both models the forecast geometry is set by the context frames.

\paragraph{The learned channel stays weak.} A small forecaster trained on retimed pairs (31 drives) ignores the plan given as a vector (plan slope .05) but uses it given as a depth map aligned with its output (.79; App.~\ref{sec:method}), a simplicity bias~\cite{shah2020pitfalls} that counterfactual pairs alone~\cite{coco2026} do not remove. DrivingWorld's camera, but not its scene, also copies a retimed history (App. Table~\ref{tab:copycat}).\label{sec:predictor}

\paragraph{Repairing the context at inference trades response for noise.} If the context sets the geometry, a repair must act on the context. Forward-warping the previous context frame along the plan at every step triples the slow-down response on held-out contexts (.18 to .54) and nearly triples its noise; seventeen variants lie on one response--noise trade-off, because local errors compound along the warp chain (App.~\ref{app:repair}). Re-sampling a logged-speed rollout (frame $\mathrm{round}(sj)$ as step $j$) lies off this trade-off, at a similar response with less than half the noise. The diagnosis sets two constraints: the plan must reach the forecast as geometry, and the generator's output must not serve as the next warp source.

%% file: sec4_retime.tex
\section{ReTime-Compose: delivering the plan as geometry}
\label{sec:compose}
ReTime-Compose takes the history and the candidate plans and returns one forecast per plan; nothing is trained. It runs in three stages (Fig.~\ref{fig:compose}, Alg.~\ref{alg:rc}): (a) the frozen generator rolls out once at the logged speed; (b) depth and camera poses read from this rollout move its frames to the cameras of each plan, the only place where the plan's geometry enters; (c) the frozen generator fills only what no rollout frame covers. Three constraints of comparing plans at decision time separate it from transport-then-complete methods for one target view~\cite{gen3c,cfwm2026,worldinworld}. \emph{The source must exist when the plan is chosen and serve every plan:} a speed plan retimes one path, so the model's own logged-speed rollout passes through all speed plans' poses. \emph{Transport must beat re-sampling that rollout,} and how much room it has depends on how far the planned poses lie from the rollout relative to the pose error of its reading (Eq.~\ref{eq:rho}). \emph{The target is the paired change:} all plans, the logged-speed one included, are processed identically. Writing each plan's forecast depth as $\widehat d_s=d_s+c+e_s$, with $c$ shared by the plans, gives $\Delta\widehat d-\Delta d=e_p-e_q$, so an error identical in all plans cancels from $\alpha$ and $N$, and only an error specific to one plan must be removed. Each step below answers a diagnosed failure; undoing it costs noise, response or image quality (Table~\ref{tab:ablmain}).
\begin{algorithm}[t]
\caption{ReTime-Compose (no training). Numbers refer to the steps of Sec.~\ref{sec:compose}.}
\label{alg:rc}
\small
\begin{algorithmic}[1]
\Require history $H$; plans $s$ with camera poses $P_{sj}$ (logged speed included); frozen generator $G$, reader $R$, refiner $F$
\State $V_1\gets G(H,\text{logged speed})$ \Comment{(1) one rollout}
\State $(D_k,\hat T_k)\gets R(V_1)$ for every rollout frame $k$ \Comment{(2) read}
\For{every plan $s$ and step $j$}
\State $\mathcal K\gets$ five rollout frames with $\hat T_k$ nearest $P_{sj}$
\State $C_{sj}\gets$ splat $\mathcal K$ to $P_{sj}$ with $D_k$ \Comment{(3) transport}
\State agents in $C_{sj}\gets$ rollout frame at physical time $j$ \Comment{(4)}
\EndFor
\For{every plan $s$}
\State $\widehat V_s\gets G(H,s)$, pin covered tokens \Comment{(5) complete}
\State paste covered pixels of $C_s$; $\widehat V_s\gets F(\widehat V_s)$
\EndFor
\State \Return $\{\widehat V_s\}$
\end{algorithmic}
\end{algorithm}

\paragraph{(1) Single-rollout source pool.} \emph{Failure:} a generator chained on its own warped frames compounds error (Sec.~\ref{sec:channel}). \emph{Fix:} the frozen generator rolls out $V_1$ once, from the real history at the logged speed, and every plan draws from this rollout; no warped frame is fed back. We use Epona~\cite{epona}, which executes the commanded speed with a gain of about 1.17, so one 36-frame rollout covers the poses of the $0.5\times$ to $1.5\times$ plans.
\paragraph{(2) Read the camera from the rollout.} \emph{Failure:} the generated camera deviates from the command (Table~\ref{tab:twolayer}). \emph{Fix:} a metric reader (VGGT~\cite{vggt}, its scale curve fitted once on exploration videos) gives per-frame depth $D_k$ and camera poses $\hat T_k$ estimated from the rollout (Appendix~\ref{app:compose}).
\paragraph{(3) Transport to the planned cameras.} \emph{Failure:} the plan tokens carry little of the geometry (Sec.~\ref{sec:channel}). \emph{Fix:} the plan's geometry is imposed only here. For step $j$ of plan $s$, the five rollout frames whose estimated poses are nearest the planned pose $P_{sj}$ are forward-splatted to the planned camera with their depth, $X_p{=}R_{pi}X_i{+}t_{pi}$, $u_p{=}\pi(K_pX_p)$; each pixel keeps the median-depth surface over its sources and the colour of the nearest source that agrees with it (depth within 10\%; after~\cite{cfwm2026}). Uncovered pixels stay uncovered. Nearby sources keep the error common: a source transported over $\delta$ is magnified by about $1{+}\delta/Z$, differently in each plan, so even the real last context frame, placed first, doubles the noise.
\begin{table}[t]
\centering\small\setlength{\tabcolsep}{3pt}
\caption{\textbf{ReTime-Compose applies a lateral plan that released models ignore.} Slope of predicted image motion on real stereo disparity for a 0.537\,m right shift (real: 1.00); two seeds; null: between seeds. $^\dagger$One seed (Orbis~2 rolls out one). $^{\ddagger}$First read. $^{\S}$Against LiDAR parallax (no stereo).}
\label{tab:lateral}
\fitcol{\begin{tabular}{lcc}
\toprule
 & slope & null (seed change)\\
\midrule
released models (Table~\ref{tab:twolayer}) & $-.30$ to .23 & --\\
\midrule
\multicolumn{3}{l}{\emph{ReTime-Compose} (Epona unless noted)}\\
Exploration (32) & .96 / .96 & .23\\
Validation (30) & .93 / .94 & .40\\
Test 1 (29) & 1.01 / 1.02 & .40\\
Test 2 (32) & .96 / .96 & .45\\
Tests 3, 4 (22, 19)$^{\ddagger}$ & .96 / .96, .99 / 1.04 & .28, .27\\
Test 5 (80)$^{\ddagger}$ & .96 / .97 & .26\\
nuScenes, Argoverse~2 (128, 26)$^{\S}$ & .95 / .94, .94 / .96 & $-.01$, .55\\
GEM, exploration (32) & .96 / 1.01 & .30\\
Orbis~2, exploration (32)$^{\dagger}$ & .90 & --\\
\bottomrule
\end{tabular}}
\end{table}
\paragraph{(4) Other vehicles keep real time.} \emph{Failure:} retiming whole frames would also retime other agents, whose motion an ego plan does not set. \emph{Fix:} static content follows the plan's clock (the retimed rollout), other vehicles physical time: detected agents are taken from the rollout frame at the physical time of step $j$ and transported with their own depth.
\paragraph{(5) Complete only the unseen.} \emph{Failure:} a generator conditioned on the plan reproduces its weak response (Sec.~\ref{sec:channel}), so it must not overwrite what transport supplies. \emph{Fix:} the generator produces each frame under the plan's actions, but during flow sampling every latent token whose pixels are mostly covered by the composite is pinned to the correspondingly noised latent of the composite, $x_t=t\,x_0+(1-t)\,\epsilon$ (RePaint~\cite{repaint} for rectified flow, the hard replacement that training-free camera control also builds on~\cite{gibbslens}); covered pixels are pasted back (without both, the slow-down response falls by .20; Sec.~\ref{sec:exp}). A single-step refiner (Difix3D+~\cite{difix3d}, released weights) then reduces splat and seam artefacts, guided by the rollout frame nearest in time, which shows the same generated world.

\paragraph{When transport beats re-sampling.} Re-sampling shows the rollout frame nearest the planned pose, displaced from it by $\Delta$; transport reaches the planned pose up to the pose error $\varepsilon$ of the reading. A static point at depth $Z$ is misplaced by about $f\delta/Z$ in the image, with $\delta{=}\Delta$ for re-sampling and $\delta{\ge}\varepsilon$ for transport, since splatting, depth and completion add error. The ratio
\begin{equation}
\rho=\Delta/\varepsilon
\label{eq:rho}
\end{equation}
compares the two misplacements from poses alone, before any forecast is read: well below one, transport adds more pose error than it removes; near or above one, it can place the scene better (App.~\ref{app:compose}). For a lateral plan $\Delta$ is the whole shift, since every rollout frame lies on the logged path.

%% file: sec5_experiments.tex
\section{Experiments}
\label{sec:exp}
\paragraph{Protocol.} For the lateral plan we regress predicted image motion on real stereo disparity (Sec.~\ref{sec:refs}); for speed plans we read the change from the logged-speed plan to the $0.5\times$ and $1.5\times$ plans against retimed LiDAR (Eq.~\ref{eq:direct-response}), two sampling seeds pooled. The baseline, \emph{re-sampling}, shows the rollout frame whose camera, read with the same pose estimates, is nearest the planned camera; the \emph{composite} is the transported frame of steps (1)--(4) before completion. Components were developed on 32 exploration contexts; we report the final configuration on a validation set and five test sets (212 contexts, 43 drives).

\subsection{Lateral plans}
\label{sec:lateral}
ReTime-Compose applies a lateral plan that the released models' conditioning ignores and re-sampling cannot follow (Table~\ref{tab:lateral}): it reaches .93--1.04 of the real stereo response on all seven KITTI sets and on GEM~\cite{gem}, .94--.96 of the LiDAR parallax on nuScenes and Argoverse~2~\cite{av2}, and .90 on Orbis~2~\cite{orbis2}. Transport sets this slope; completing the uncovered 15\% of each frame raises it slightly (App.~\ref{app:compose}). It holds at .91--.94 up to 1.61\,m, is .63--.77 at one lane (3.5\,m) and stays at .90--.99 when the shift is combined with a speed plan (App.~\ref{app:composite}).

\begin{figure}[t]
\centering\includegraphics[width=\columnwidth]{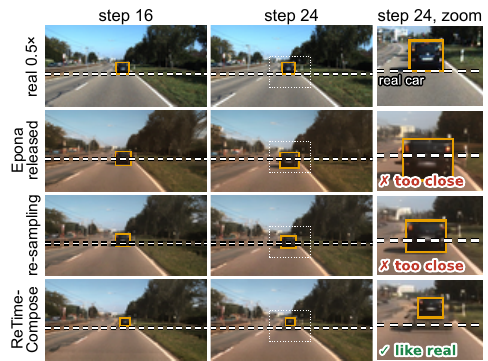}
\caption{\textbf{Under a slow-down plan, ReTime-Compose shows static content at the planned pose.} Real pose-matched video, the released model, re-sampling and ReTime-Compose at steps 16 and 24 of the $0.5\times$ plan; right: step 24 enlarged. Dashed line: the bottom of the real lead car; marks: whether the generated car ends there or closer (more: App.~Fig.~\ref{fig:qual}).}
\label{fig:qual05}
\end{figure}
\begin{table}[t]
\centering\small\setlength{\tabcolsep}{3pt}
\caption{\textbf{ReTime-Compose raises Epona's speed response over re-sampling on every set.} Contexts: validation 30; test sets 1--5: 29, 32, 22, 19, 80 ($^{*}$79). Bold: not worse than re-sampling. Upper reference: the real future transported (no refiner), unavailable at decision time.}
\label{tab:compose}
\fitcol{\begin{tabular}{llcccc}
\toprule
 & & \multicolumn{2}{c}{$0.5\times$} & \multicolumn{2}{c}{$1.5\times$}\\
Set & Method & $\alpha\uparrow$ & $N\downarrow$ & $\alpha\uparrow$ & $N\downarrow$\\
\midrule
Validation & unmodified Epona & .24 & 2.76 & .19 & 3.56\\
 & re-sampling & .24 & 1.61 & .19 & 1.96\\
 & composite, no completion & .55 & 1.51 & .39 & 2.16\\
 & \textbf{ReTime-Compose} & \textbf{.48} & \textbf{1.07} & \textbf{.33} & \textbf{1.43}\\
 & upper ref.: real future & .62 & .60 & .59 & .78\\
 & real video & .73 & .36 & .73 & .26\\
\midrule
Test 1 & re-sampling & .20 & 2.77 & .11 & 2.06\\
 & \textbf{ReTime-Compose} & \textbf{.43} & \textbf{2.30} & \textbf{.18} & \textbf{1.93}\\
Test 2 & re-sampling & .29 & 2.01 & .14 & 2.17\\
 & \textbf{ReTime-Compose} & \textbf{.34} & \textbf{1.74} & \textbf{.47} & 3.61\\
Test 3 & re-sampling & .50 & 1.90 & .28 & 2.48\\
 & \textbf{ReTime-Compose} & \textbf{.65} & \textbf{1.54} & \textbf{.42} & 3.42\\
Test 4 & re-sampling & .44 & 1.81 & .25 & 1.89\\
 & \textbf{ReTime-Compose} & \textbf{.57} & \textbf{1.36} & \textbf{.36} & \textbf{1.46}\\
Test 5 & unmodified Epona$^{*}$ & .30 & 2.27 & .20 & 2.94\\
 & re-sampling & .27 & 1.74 & .13 & 2.59\\
 & \textbf{ReTime-Compose} & \textbf{.41} & \textbf{1.47} & \textbf{.24} & \textbf{2.12}\\
\bottomrule
\end{tabular}}
\end{table}

\subsection{Speed plans}
\label{sec:speed}
\begin{figure}[t]
\centering\includegraphics[width=\columnwidth]{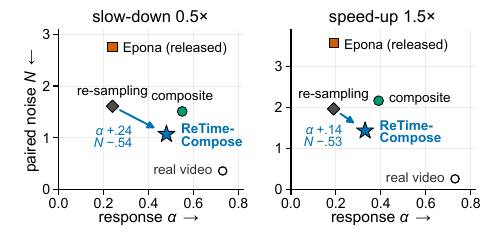}
\caption{\textbf{Completion takes ReTime-Compose off the response--noise trade-off} (validation). Re-sampling lowers the noise, the composite raises the response, and completion brings the noise below re-sampling's at both speeds (arrow: the change from re-sampling).}
\label{fig:frontier}
\end{figure}
\paragraph{The speed response rises on every set.} Over re-sampling (Table~\ref{tab:compose}, Fig.~\ref{fig:qual05}), ReTime-Compose raises it by $+.17$ and $+.16$ pooled over the validation set and test sets 1--4, and by $+.14$ and $+.11$ on test set 5, 80 contexts from 25 further drives read once. On GEM and Orbis~2 (96 contexts) the slow-down response rises by $+.24$ and $+.16$, intervals above zero. Transport and the main reading both use VGGT, so we re-read with an independent reader (Depth Anything V2~\cite{da2}: $+.12$ to $+.39$ and $+.22$ to $+.39$ over validation and test sets 1--4) and with no depth reader, from optical flow against LiDAR ($+.29$, interval incl.\ zero, and $+.35$); the gain holds under the others and four further controls (App. Table~\ref{tab:controls}) and for braking and stopping plans (App.~\ref{app:timevar}). Most of the remaining gap to real video comes from the rollout's geometry, not from transport: the real future transported reaches .62/.59 (real video .73). The noise is below re-sampling's at the slower speed and, without a depth reader, in total error at both speeds (Fig.~\ref{fig:frontier}; App.~\ref{app:t5}).

\paragraph{Eq.~\ref{eq:rho} predicts where transport pays off.} In every case of App.~Table~\ref{tab:rho} (generator, plan, data set), no case with $\rho\le.39$ shows a lower error and every lower error has $\rho\ge.9$; transport gains most on the lateral plan.

\subsection{Comparison and ablations}
\label{sec:cmp}
\paragraph{Closest approaches.} Transporting real frames of the target trajectory~\cite{cfwm2026,worldinworld} with our completion is the upper reference of Table~\ref{tab:compose}; from the rollout alone ReTime-Compose closes 35--63\% of the gap from re-sampling to it. Feeding warped frames through the history path (Warp-as-History~\cite{warpashistory} without its one-video LoRA) gives a similar response at 3.6--3.7$\times$ the noise, as Sec.~\ref{sec:channel} predicts. Without a world model, rigidly transporting the last context frame in RGB responds (.66/.41) at 7--12$\times$ our noise (App.~\ref{app:compose}). GEN3C~\cite{gen3c}, trained for camera control through a 3D cache rendered from the last frame, responds less than ReTime-Compose given the same plan cameras (.39 and .13 against .49 and .27 on 11 validation contexts; App.~\ref{app:compose}).
\begin{table}[t]
\centering\small\setlength{\tabcolsep}{3pt}
\caption{\textbf{Undoing a step costs response or noise.} Ablated minus full method at $0.5\times$ / $1.5\times$; negative $\Delta\alpha$ and positive $\Delta N$ are worse; bold: 95\% drive-cluster interval excludes zero. Epona validation set unless noted (full table with intervals: App.~Table~\ref{tab:ablation}).}
\label{tab:ablmain}
\begin{tabular}{lll}
\toprule
Change & $\Delta\alpha$ & $\Delta N$\\
\midrule
camera from the command & $-$.14 / $-$.14 & \textbf{+.64} / \textbf{+.92}\\
\quad GEM, test sets 3--4 (41) & \textbf{$-$.27} / -- & $-$.22 / --\\
1 source, not 5 & $-$.02 / +.02 & \textbf{+.19} / \textbf{+.36}\\
last real frame placed first & $-$.01 / +.06 & \textbf{+1.28} / \textbf{+2.24}\\
other vehicles on the plan clock & +.01 / +.02 & +.00 / \textbf{+.21}\\
no completion (vs.\ unrefined) & +.13 / +.05 & \textbf{+.50} / \textbf{+.73}\\
neither pinning nor paste-back & \textbf{$-$.20} / $-$.08 & \textbf{+1.86} / \textbf{+1.27}\\
\quad test set 3 (22) & $-$.18 / \textbf{$-$.32} & +.72 / +3.63\\
\quad test set 5 (80) & \textbf{$-$.09} / $-$.06 & \textbf{+.88} / \textbf{+1.26}\\
Orbis~2, no pinning & \textbf{$-$.14} / \textbf{$-$.11} & +.28 / \textbf{+.66}\\
\bottomrule
\end{tabular}
\end{table}

\paragraph{Each step pays for itself.} Reading the camera instead of taking the command (Table~\ref{tab:ablmain}) keeps .14 of response on Epona and .27 on GEM; nearby sources, the physical clock for other vehicles and completion keep the noise down. Letting the generator own the covered pixels (neither pinning nor paste-back) brings back the diagnosed failure: the slow-down response falls by .20 $[.09,.36]$, also on test sets 3 and 5, read after the method was frozen, and on Orbis~2.

\subsection{Image quality}
\label{sec:quality}
Against real frames, colour from the nearest agreeing source gives KID~\cite{kid} .021--.030 on the validation and test sets 1--4 (rollout .014--.016), and the refiner lowers it by 22--40\%. On DINOv2 ViT-L/14 features~\cite{dinov2,stein2023exposing} Fr\'echet distance is below re-sampling's on all five.

%% file: sec6_limits.tex
\section{Limitations and conclusion}
\label{sec:limits}
\paragraph{Scope.} Retiming covers static content under speed plans; stereo and LiDAR parallax cover lateral shifts up to 3.5\,m; neither covers turns, lane-change manoeuvres or the reactions of other agents, kept at physical time. ReTime-Compose needs a rollout that reaches the planned poses (App.~Table~\ref{tab:where}; GEM's: slow-down only) and a pose reading that works on generated frames. It changes where the frozen generator's content goes, not what it draws: its forecasts are as sharp as the generator's own logged-speed rollout. A context (three plans, two seeds) takes under 12\,min on one RTX 4080 SUPER. Its noise gain depends on $\rho$ (Eq.~\ref{eq:rho}; below one on nuScenes and Orbis~2). Closed-loop planning with these forecasts remains to be tested.

\paragraph{Conclusion.} A generated trajectory is not evidence that the scene follows the plan. Real references for undriven plans show that released world models move the camera but not the scene, because the plan does not reach the forecast geometry; delivered as geometry, it is followed without training. Both references come from existing driving logs, so a world model can be checked against them before it compares plans. References and code will be public.

%% file: supp_compose.tex
\section{Context repair inside the generator: details}
\label{app:repair}
\paragraph{Repairing through the context.} The diagnosis fixes where a training-free repair must act. A learned per-pixel hint on the plan tokens fails in free generation ($\alpha{=}.09/{-}.05$) but works under real history (.36/.58): it conflicts with the model's own context. We therefore change the context: at every generated step the previous context frame is forward-warped by the plan's one-step motion with carried metric depth, the covered tokens (${\ge}95\%$ of their pixels) replace the latest context frame, and holes keep the model's own content; output frames are not anchored and nothing is trained. This is the render--repeat structure of 3D-cache conditioning~\cite{gen3c}; what the diagnosis adds is \emph{why} the context and not the token is the place to act. On the 32 exploration contexts the step-wise warp raises $\alpha$ from .19/.09 to .37/.20, and estimating the depth of hole content with a monocular metric model raises it to .54/.26. On a confirmation set frozen before any method reading (32 contexts, 12 drives, never used for tuning), one run of this configuration gives .54/.30 against .18/$-$.08 for the unmodified model, with disjoint drive-cluster intervals.

\paragraph{The price, and where it comes from.} The response gain is paid in paired noise $N{=}(1{-}\mathrm{EF})-(1{-}\alpha)^2$, the residual variance of the predicted change orthogonal to the true change (1.06/1.39 unmodified, 4.26/4.90 with hole depth). Seventeen variants of gating, encoding, sampling and rendering lie on one $\alpha$--$N$ trade-off, whereas real history reaches $\alpha{\approx}.6$ at $N{<}1$. Six explanations stated in advance (gating power, context encoding, sampling, hole content, shared sources, resampling order) were each rejected (Appendix~\ref{app:noise}). Splitting the paired residual into the two plans' own depth errors, $\Delta\hat d-\Delta d=e_p-e_q$, locates it: after removing each frame's global scale, real-history oracles carry local errors of 0.3--1.0 (relative to the true change) and the unmodified model 1.0--1.6, while every plan whose context is warped from its own previous composite carries 3.7--5.8, including the logged-speed reference plan. Errors compound along the warp chain because the model's own generation never re-enters it; Epona shows the same pattern. Refining from the model's own frame instead lowers the reference plan's local error (4.2$\to$2.7) but halves the response on both generators: the generated frame reverts toward the default motion, so the plan state carried by the chain is lost. On the confirmation set the paired noise rises from .65/1.07 to 1.73/3.16 (Appendix~\ref{app:noise}); the finding-to-design chain: App.~\ref{app:chain}. A training-free baseline that never touches the generator shows how much of this noise is avoidable: re-sampling the logged-speed rollout, taking frame $\mathrm{round}(sj)$ as step $j$ of plan $s$, gives $\alpha{=}.47$ at $N{=}1.79$ for the slow-down on the same contexts, against .54 at 4.26 for the best context repair.

\section{ReTime-Compose: pose reading, ablations}
\label{app:compose}

\paragraph{Where ReTime-Compose applies.} Table~\ref{tab:where} collects every generator, data set and plan we read. Two conditions decide the outcome, both checkable from poses before any forecast is read: the rollout must reach the planned positions along the path, a lateral plan being a sideways offset of them (otherwise there is nothing to transport), and for speed plans the offset of the nearest rollout frame must not be small against the pose error of the reading ($\rho$, Eq.~\ref{eq:rho}). A lateral plan always satisfies the second, since no rollout frame is shifted.

\paragraph{Completion operator.} Swapping only the completion operator of step (5) for NVS-Solver's blending of the score with warped views~\cite{nvssolver} gives the same alignment and noise on the validation set at a higher KID (.031 vs .025).

\begin{table}[h]
\centering\small\setlength{\tabcolsep}{2.5pt}
\caption{\textbf{Where ReTime-Compose applies.} Rollout: whether the frozen rollout reaches the planned positions along the path (lateral: offset sideways from them). Result: response of ReTime-Compose (lateral and composite: slope against real stereo or LiDAR parallax, real 1.00; speed: gain in $\alpha$ over re-sampling). n/a: the rollout does not reach the planned poses.}
\label{tab:where}
\fitcol{\begin{tabular}{lllccl}
\toprule
Generator & Data & Plan & Rollout & $\rho$ & Result\\
\midrule
Epona & KITTI & $0.5\times$ / $1.5\times$ & yes & 1.00 / 2.67 & $+.17$ / $+.16$ (5 sets)\\
Epona & KITTI & lateral 0.27--1.61\,m & yes & -- & .91--1.04\\
Epona & KITTI & lateral 3.5\,m & yes & -- & .63--.77\\
Epona & KITTI & $0.5\times$ / $1.5\times$ + 0.54\,m & yes & -- & .96--.99 / .90--.96\\
Epona & KITTI & brake / stop & yes & -- & $+.15$ / $+.10$\\
Epona & nuScenes & $0.5\times$ & yes & .34 & response up, noise up\\
Epona & nuScenes & lateral 0.54 / 1.61 / 3.5\,m & yes & -- & .95 / .96 / .63\\
Epona & nuScenes & $0.5\times$ + 0.54\,m & yes & -- & .99 / .99\\
Epona & Argoverse~2 & lateral 0.54\,m & yes & -- & .94 / .96\\
Epona & Argoverse~2 & speed & no & -- & n/a\\
GEM & KITTI & $0.5\times$ & yes & .90 & $+.24$\\
GEM & KITTI & $1.5\times$ & no & -- & n/a\\
GEM & KITTI & lateral 0.54\,m & yes & -- & .96 / 1.01\\
Orbis~2 & KITTI & $0.5\times$ & yes & .39 & $+.16$, noise not lower\\
Orbis~2 & KITTI & lateral 0.54\,m & yes & -- & .90\\
\bottomrule
\end{tabular}}
\end{table}

\begin{figure*}[t]
\centering\includegraphics[width=.84\textwidth]{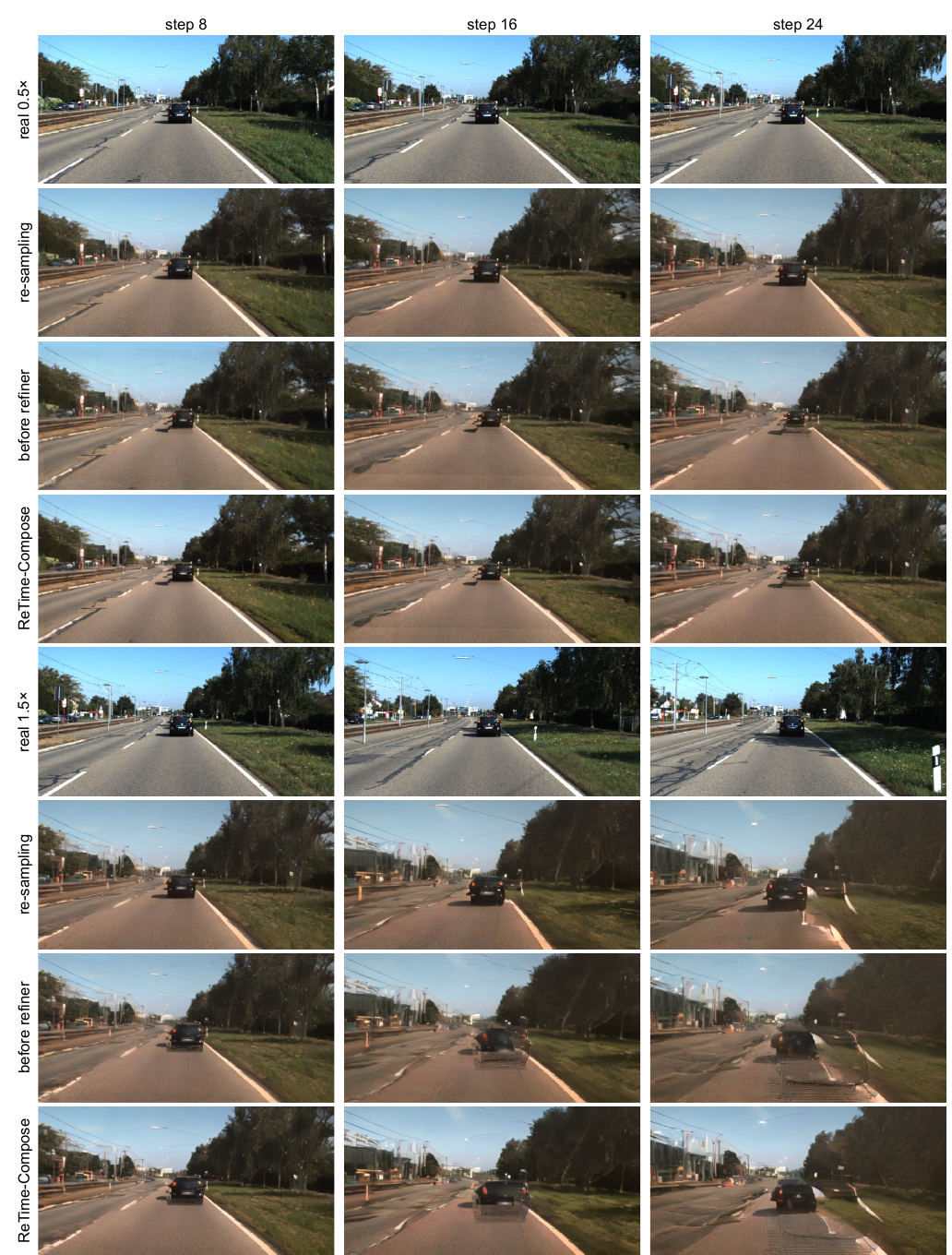}
\caption{\textbf{Speed plans, one exploration context} (chosen as the context whose slow-down gain over re-sampling is closest to the set's median). Rows: real pose-matched video, re-sampling the rollout, ReTime-Compose before the refiner, and the final method, for the $0.5\times$ and $1.5\times$ plans at steps 8, 16, 24. The composite places static content at the planned pose; the refiner reduces the splat artefacts the completion would otherwise reproduce (image quality: Sec.~\ref{sec:quality}).}
\label{fig:qual}
\end{figure*}

\paragraph{Pose reading.} A metric reader gives per-frame depth $D_k$ (VGGT~\cite{vggt}, its scale drift along the clip calibrated once on real video). Camera poses $\hat T_k$ are \emph{estimated} from the generated frames: camera centres come from a frame-to-frame RAFT~\cite{raft}+PnP chain on the read depth, rotations from a three-frame keyframe chain with fallback to the frame-to-frame rotation where the two disagree by more than $1^\circ$. On real video this reading gives a median forward offset of $-.14$\,m and a median absolute yaw error of $.27^\circ$ after 24 frames; on generated frames we require it to agree with an independent VGGT pose chain in path length within 20\%.

\paragraph{Refiner.} Step~(5) runs the released reference-conditioned Difix3D+ model~\cite{difix3d} (one denoising step at timestep 199, prompt ``remove degradation'', no guidance, nothing trained) on frames 1--24 of every arm at the evaluation resolution ($576{\times}1024$); frame 0, the real context, is unchanged. The reference for step $j$ of plan $s$ is the logged-speed rollout frame $\mathrm{round}(sj/k)$, with $k$ the generator's executed-speed gain (1.17 Epona, .83 GEM, 1.34 Orbis~2), i.e.\ the rollout frame nearest in time to the planned pose. This reference and the settings were chosen on the validation set only. On one RTX 4080 SUPER (about 15\,GB) it takes about 3.8 minutes for a context's six 24-frame arms (about 1.6\,s per frame).

\paragraph{Ego-lane road surface.} On the ego-lane road surface, where re-sampling barely responds (.24/.07; real .69/.67), ReTime-Compose reaches .63/.39.

\paragraph{Moving objects.} Retiming references static content; on KITTI, moving objects enter all arms alike. Dropping the supported points that fall in a detected person, bicycle, car, motorcycle, bus or truck box (score ${\ge}.5$) in either matched real frame (6.8\% of the points; validation and test sets 1--3, 113 contexts) changes the real-video response by at most .03 (.72/.70 to .69/.68) and the gain over re-sampling from $+.17$/$+.17$ to $+.20$ $[+.14,+.28]$ and $+.20$ $[+.12,+.35]$.

\paragraph{Re-sampling with the same refiner.} Because the refiner runs on our output, a fair baseline passes re-sampling through the same model, reference rule and settings. On re-sampling it leaves the response unchanged ($\Delta\alpha$ within $\pm.03$ on every set and speed). Against this baseline, pooled over the five evaluation sets (132 contexts, 18 drives, the contexts and seeds of Table~\ref{tab:compose}), ReTime-Compose raises the response by $+.16$ $[+.10,+.22]$ at $0.5\times$ and $+.17$ $[+.09,+.27]$ at $1.5\times$ and lowers the slow-down total error by $.75$ $[.47,1.08]$, the same pattern as against re-sampling (Table~\ref{tab:compose}): the gain does not come from the refiner.

\begin{table*}[t]
\centering\footnotesize\setlength{\tabcolsep}{4pt}
\caption{\textbf{Component and joint-safeguard ablations} (final configuration; Epona validation set, 30 contexts; Orbis~2 validation set, 32 contexts). Change in response $\alpha$ and noise $N$ at $0.5\times$ / $1.5\times$, ablated minus full method, 95\% drive-cluster intervals where they exclude zero; negative $\Delta\alpha$ and positive $\Delta N$ are worse. Each row undoes one design choice (``neither'': both protections of step 5); the second column names the failure that choice answers. $^\dagger$GEM test sets 3 and 4 (41 contexts, 9 drives), slow-down only (GEM's scored plan). $^\ddagger$Paired with the frozen method on the same set: test set 3, 22 contexts, 8 drives; test set 5, 80 contexts, 25 drives. $^{\S}$The composite is unrefined, so it is paired with the full method before the refiner ($\alpha$/$N$ .42/1.01 and .34/1.43); every other Epona row is paired with the refined output of Table~\ref{tab:compose}.}
\label{tab:ablation}
\begin{tabular}{lllc}
\toprule
Change & Failure it answers & $\Delta\alpha$ & $\Delta N$\\
\midrule
camera from the command, not read & executed $\neq$ commanded & $-$.14 / $-$.14 & +.64 [.14,1.80] / +.92 [.39,1.81]\\
1 source, not 5 & per-source geometry error & $-$.02 / +.02 & +.19 [.01,.50] / +.36 [.11,.82]\\
last real frame placed first & arm-specific magnification & $-$.01 / +.06 & +1.28 [.52,2.88] / +2.24 [1.07,4.64]\\
other vehicles on the plan clock & agents keep physical time & +.01 / +.02 & +.00 / +.21 [.04,.50]\\
no completion$^{\S}$ & generated holes & +.13 / +.05 & +.50 [.24,.91] / +.73 [.20,1.70]\\
no pinning & \multirow{5}{*}{generator ignores the plan} & $-$.02 / $-$.01 & $-$.05 / +.34 [.01,.90]\\
no paste-back & & +.00 / +.03 & +.08 / +.30 [.08,.66]\\
neither & & $-$.20 [$-$.36,$-$.09] / $-$.08 & +1.86 [.60,4.84] / +1.27 [.45,3.07]\\
neither, test set 3$^\ddagger$ & & $-$.18 / $-$.32 [$-$.49,$-$.16] & +.72 / +3.63\\
neither, test set 5$^\ddagger$ & & $-$.09 [$-$.20,$-$.02] / $-$.06 & +.88 [.32,2.03] / +1.26 [.48,2.45]\\
no refiner & splat artefacts (quality) & $-$.06 / +.01 & $-$.06 / $-$.01\\
\midrule
Orbis~2, no pinning & generator ignores the plan & $-$.14 [$-$.27,$-$.02] / $-$.11 [$-$.19,$-$.00] & +.28 / +.66 [.14,1.30]\\
GEM, camera from the command$^\dagger$ & executed $\neq$ commanded & $-$.27 [$-$.41,$-$.14] / -- & $-$.22 / --\\
\bottomrule
\end{tabular}
\end{table*}

\paragraph{Lateral intervals.} On test sets 3 and 4 (Table~\ref{tab:lateral}, first reads with the frozen method) the lateral slope is .96 $[.92,1.00]$ and .96 $[.91,1.00]$ (test set 3, two seeds) and .99 $[.90,1.04]$ and 1.04 $[.98,1.09]$ (test set 4), 95\% drive-cluster intervals; real stereo reads .99 on both.

\paragraph{Lateral magnitude.} Stereo fixes the shift at the baseline. To test other magnitudes we scale the planned shift to $r B$ with $r\in\{.5,2,3\}$ (0.27, 1.07 and 1.61\,m) and to 3.5\,m, about one lane, and take the reference image motion from LiDAR parallax, $-f r B/Z$, since no real camera sits there (criterion written before each reading: interval above zero). On the validation set (Epona, 30 contexts, two seeds) the slope is .93 / .93 at 0.27\,m, .93 / .94 at 0.54\,m, .92 / .94 at 1.07\,m ($[.89,.96]$, $[.90,.97]$), .91 / .93 at 1.61\,m ($[.88,.95]$, $[.89,.96]$) and .76 / .77 at 3.5\,m ($[.73,.80]$, $[.73,.82]$); the null between seeds falls from .40 at 0.54\,m to .14 at 1.61\,m and .06 at 3.5\,m. The 1.61\,m arm was read against the 0.54\,m reference and divided by $r{=}3$, which is exact because the slope is linear in the reference. The predicted image displacement stays proportional to the planned shift up to 1.61\,m; at one lane the response falls to about three quarters, as the rollout stays in its lane and completion fills larger disoccluded regions. Transport alone (the composite without completion, same 30 validation contexts, read once against a description written before) gives .92 / .94 at 0.54\,m (null .39) with 15\% of each frame uncovered, and .71 / .71 at 3.5\,m (null .06) with 42\% uncovered: the slope is set by transport, and at one lane completion raises it to .76 / .77. Test set 5 (80 contexts, 25 drives, read once) repeats this: transport alone gives .70 / .71 at 3.5\,m with 46\% uncovered, completion raises it to .73 / .74 ($+.03$ $[+.02,+.04]$ and $+.04$ $[+.03,+.04]$, paired by context), and the noise is unchanged. On KITTI the slopes of Table~\ref{tab:lateral} are read before the refiner; with it they move by at most .01. On the five test sets the 3.5\,m slopes are .70 $[.66,.79]$ / .68 $[.62,.78]$ (test set 1), .70 $[.66,.74]$ / .71 $[.67,.75]$ (2), .63 $[.50,.76]$ / .63 $[.52,.75]$ (3), .69 $[.54,.76]$ / .73 $[.60,.81]$ (4) and .73 $[.68,.78]$ / .74 $[.70,.79]$ (5), and on nuScenes (against LiDAR parallax) .63 $[.60,.66]$ / .62 $[.58,.65]$ (128 contexts, null $-.00$), after .96 $[.89,1.04]$ / .93 $[.86,1.02]$ at 1.61\,m (32 contexts, null $-.03$), each read once against a criterion written before it.

\paragraph{Composite plan.}\label{app:composite} The plan retimes the logged path to $1.5\times$ (or $0.5\times$) and shifts it to the right by $d(j){=}B\min(1,j/10)$, the ramp of the lateral plan. At even steps $j\ge10$ the retimed camera is the logged left camera of frame $1.5j$ (or $0.5j$), so the composite camera is the logged right camera of that frame and the real stereo pair remains the reference. Built from real video (two exploration contexts), the composite reads 1.03 against it (real pair: 1.00). On the validation set and test set 1 (59 contexts, 17 drives; criterion written before the single read: both lower bounds above zero) ReTime-Compose reads .90 $[.85,.95]$ and .96 $[.88,1.03]$ (real .99). The null between the two seeds is .78 $[.43,1.25]$, against .23--.45 at $1\times$ (Table~\ref{tab:lateral}), in line with the larger share of completion at $1.5\times$, where the planned cameras run furthest beyond the rollout. With the slow-down instead (same 59 contexts, criterion written before the read), ReTime-Compose reads .96 $[.93,.98]$ and .96 $[.92,.99]$ (real .99), null .21. On test set 5 (80 contexts, 25 drives, read once) the speed-up composite reads .90 $[.85,.96]$ and .93 $[.87,.99]$ (real .99), null .45, and the slow-down composite .99 $[.95,1.03]$ and .97 $[.94,.99]$, null .20. On nuScenes (48 contexts, 18 drives; reference $-f\,d(j)/Z$ from static LiDAR points moved into the $0.5\times$ plan camera, read once against a criterion written before it) the slow-down composite reads .99 $[.95,1.03]$ and .99 $[.95,1.05]$.

\begin{table}[h]
\centering\small\setlength{\tabcolsep}{3pt}
\caption{\textbf{The gain over re-sampling persists under each control.} Change of $\alpha$ over re-sampling, pooled, with 95\% drive-cluster intervals; second reader: range over the validation set and test sets 1--4.}
\label{tab:controls}
\fitcol{\begin{tabular}{lccc}
\toprule
Control & Contexts & $0.5\times$ & $1.5\times$\\
\midrule
none (main reader) & 132 & $+.17$ $[+.10,+.23]$ & $+.16$ $[+.08,+.25]$\\
second depth reader & 132 & $+.12$ to $+.39$ & $+.22$ to $+.39$\\
no depth reader (flow) & 87, 89 & $+.29$ $[-.05,+.83]$ & $+.35$ $[+.10,+.48]$\\
re-sampling also refined & 132 & $+.16$ $[+.10,+.22]$ & $+.17$ $[+.09,+.27]$\\
people, vehicles removed & 113 & $+.20$ $[+.14,+.28]$ & $+.20$ $[+.12,+.35]$\\
non-overlapping windows & 82 & $+.19$ $[+.10,+.27]$ & $+.16$ $[+.05,+.28]$\\
static obstacle points & 123--125 & $+.14$ $[+.01,+.24]$ & $+.18$ $[+.05,+.32]$\\
\bottomrule
\end{tabular}}
\end{table}

\paragraph{Time-varying plans.}\label{app:timevar} Retiming covers any monotone time warp of the logged path. We read two that start at the logged speed: BRAKE, $\tau(j)=j-j^2/64$ (speed ratio from 1 to .25 over 2.4\,s), and STOP, $\tau(j)=j-j^2/32$ up to $j=16$ and constant after (standstill from 1.6\,s). At steps 16 and 24 both $\tau$ and the logged-speed time are integers, so the recorded frame and LiDAR are the reference. On test set 5 (80 contexts, 25 drives, both seeds pooled, read once against a criterion written before the runs), ReTime-Compose reads .36 (BRAKE) and .41 (STOP) against .21 and .31 for re-sampling (real video .79 and .81): $\Delta\alpha$ $+.15$ $[+.08,+.25]$ and $+.10$ $[+.04,+.20]$, with the noise lower by .58 $[.06,1.73]$ and .37 $[.02,1.17]$. Only the plan codes change; the frozen chain is otherwise the one of Table~\ref{tab:compose}.

\paragraph{Test set 5.}\label{app:t5} Built after the method was frozen from 25 KITTI drives that no other set uses, with a fixed selection rule (free and lead windows, at least 50 frames apart, at most eight per drive). Four windows whose logged motion the frozen generator's own action conversion rejects were excluded before any output, leaving 80 contexts; the criterion (both lower interval bounds of $\Delta\alpha$ above zero) was written before the single read. ReTime-Compose reads $\alpha$ .41 / $N$ 1.47 at $0.5\times$ and .24 / 2.12 at $1.5\times$; re-sampling .27 / 1.74 and .13 / 2.59; re-sampling passed through the same refiner .29 / 1.78 and .13 / 2.76; the released model conditioned on the same plans .30 / 2.27 and .20 / 2.94 (79 contexts: one more window fails its action conversion at $0.5\times$). Against refined re-sampling $\Delta\alpha$ is $+.12$ $[+.07,+.22]$ and $+.11$ $[+.05,+.20]$, $\Delta N$ $-.31$ $[-.92,+.08]$ and $-.64$ $[-1.61,-.11]$. Without a depth reader (criterion written in advance: both $P(\Delta\alpha>0)\ge.90$) the gain is $+.14$ $[-.01,+.25]$ ($P$ .97) at $1.5\times$ and $+.11$ $[-.08,+.26]$ ($P$ .82) at $0.5\times$, so this check passes at the faster speed only; re-sampling already reaches .81 of the real .99 at $0.5\times$ under this measure, which leaves less room.

\paragraph{Trained explicit-geometry camera control.} GEN3C~\cite{gen3c} builds a 3D cache from the MoGe~\cite{moge} depth of a single image and renders it along a given camera path. We give it the last context frame and our plan cameras (translations converted to MoGe's scale by the median MoGe/LiDAR depth ratio of that frame) and read its output with the same reader (validation set, first 11 contexts from 7 drives, one seed, as written before the run). ReTime-Compose has the higher response at both speeds (.49 against .39 and .27 against .13; differences $-.11$ $[-.31,-.00]$ and $-.13$ $[-.44,-.04]$ for GEN3C); GEN3C's noise is similar at $0.5\times$ and lower at $1.5\times$ (1.01 against 1.59), where it barely follows the plan: a cache built from one frame does not reach far ahead, and the generator fills in a smooth view that does not move as planned. Re-sampling reads .26 and .16 on the same contexts.

\paragraph{Image quality per set.} With colour from the nearest agreeing source, KID is .025 on the validation set and .030, .027, .028 and .021 on test sets 1--4; with the refiner, .020, .018, .019, .016 and .015 (Sec.~\ref{sec:quality}), and .013 on test set 5 (rollout .010); on the four sets for which unrefined outputs are available, the 95\% drive-bootstrap interval of the decrease (same frames, full-set KID) lies above zero on each: .0055 $[.0029,.0103]$ on validation, .0119, .0078 and .0109 on test sets 1--3. 

\paragraph{Non-overlapping windows and the unrefined output.} Counting only non-overlapping windows (82 of the 132 pooled contexts; windows of different sets on the same drive may share frames), $\Delta\alpha$ is $+.19$ $[+.10,+.27]$ and $+.16$ $[+.05,+.28]$ and $\Delta N$ $-.27$ $[-.57,-.02]$ and $+.25$ $[-.39,+.83]$. Without the refiner, ReTime-Compose gives $\alpha$/$N$ .42/1.01 and .34/1.43 on the validation set.

\paragraph{Orbis~2 speed plans.} The same recipe (no per-model tuning) runs on Orbis~2 with pinning inside its sampler's step schedule and the VGGT pose reading. On the same three sets as GEM (73 contexts: 32, 22 and 19; one seed) the method (with the refiner) raises the slow-down response over re-sampling by $+.23$ $[+.11,+.43]$ and the speed-up response by $+.17$ $[+.05,+.30]$; on the fourth set it does not ($-.08$ and $-.04$, $\rho$ .60 and 1.11), and over all four sets the gains are $+.16$ $[+.07,+.30]$ and $+.11$ $[+.02,+.24]$; completion lowers the noise of the composite alone ($-.28$ $[-.82,-.01]$ at $0.5\times$), but the noise is not lower than re-sampling's. Computed after this read-out, Eq.~\ref{eq:rho} with Orbis~2's VGGT pose chain gives $\rho$ .39 at the slow-down and .88 at the speed-up on test sets 3 and 4 (offsets .36 and .81\,m against a .92\,m pose error on real video), below one as on nuScenes; on GEM ($\rho$ .90 on the same sets) the depth reader finds the noise not lower either (without a depth reader: Appendix~\ref{app:flow}).

\paragraph{GEM speed plans and implementation.} GEM's logged-speed rollout covers about .83 of the logged path. Commanding the source rollout $1.5\times$ faster so that it reaches farther poses does not help: on the exploration contexts (28, 13 drives) the method's slow-down total error rises by .43 $[.02,1.26]$, consistent with Sec.~\ref{sec:diagnosis} (off the logged command the model moves its camera, not its scene). Hence only slow-down plans are scored. On three sets (validation 32, test sets 3 and 4 with 22 and 19 contexts; two sampling seeds) the method with the refiner raises the response by $+.28$ $[+.18,+.37]$ ($+.26$, $+.13$, $+.46$ per set); its noise is not lower. A fourth set (23 contexts, 8 drives of test sets 1 and 2) gives $+.10$ $[-.20,+.31]$ (not significant), with $\rho$ .68 and .91 on its two halves and gains of $-.11$ and $+.27$; over all four sets the gain is $+.24$ $[+.14,+.35]$. The refiner lowers GEM's KID from .024, .028 and .028 to .019, .019 and .022 on the three sets and leaves the response unchanged ($\Delta\alpha$ $-.02$, $-.03$, $-.02$ against the unrefined output). Under the second reader (DA2) the slow-down response gain holds on GEM ($+.25$ and $+.44$ on test sets 3 and 4) and on Orbis~2 ($+.39$ and $+.36$; speed-up $+.31$ and $+.34$), with real video at .94--1.03. On the exploration set (32 contexts, one sampling seed), the method raises the slow-down response over re-sampling ($+.18$ $[.00,+.85]$) and completion lowers its noise by 1.95 $[.15,7.81]$, but its noise stays .20 above re-sampling (not significant). One rollout and its reading serve every plan, arm and baseline; completion takes about seven minutes per context on one 24\,GB GPU.

\paragraph{What the world model adds.} Rigidly transporting the last context frame along the plan in RGB, with no world model, is unusable: on the exploration contexts it gives $\alpha$ .66/.41 at $N\approx$13/35, against .42/.34 at 1.81/2.92 for ReTime-Compose before the refiner on the same contexts and instrument. Rollout and completion cut the noise by an order of magnitude.

\begin{table}[t]
\centering\small\setlength{\tabcolsep}{3pt}
\caption{\textbf{Second reader} (Depth Anything V2~\cite{da2} for all arms; real video $\alpha$ .83--1.03); $\alpha$ / $N$, all steps.}
\label{tab:da2}
\begin{tabular}{llcc}
\toprule
Set & Arm & $0.5\times$ & $1.5\times$\\
\midrule
Validation & re-sampling & .18 / 2.76 & .23 / 3.49\\
 & ReTime-Compose & .57 / 1.54 & .50 / 2.35\\
Test 1 & re-sampling & .24 / 3.40 & .15 / 3.20\\
 & ReTime-Compose & .58 / 2.27 & .37 / 3.05\\
Test 2 & re-sampling & .62 / 5.20 & .28 / 3.56\\
 & ReTime-Compose & .80 / 4.00 & .55 / 4.26\\
Test 3 & re-sampling & .74 / 3.02 & .26 / 4.16\\
 & ReTime-Compose & .91 / 2.85 & .65 / 4.65\\
Test 4 & re-sampling & .67 / 2.99 & .40 / 3.62\\
 & ReTime-Compose & .79 / 1.42 & .65 / 1.96\\
\bottomrule
\end{tabular}
\end{table}

\paragraph{Sampling seeds.} The two completion seeds of every arm are fixed in advance. Repeating the final configuration on all five evaluation sets (132 contexts, same rollouts and reads) with completion seeds 411 and 2004 moves the pooled response by $+.01$ and $-.02$ (at most .08 on any set) and the validation KID from .0196 to .0198. The noise is the more seed-sensitive quantity: pooled $N$ is 1.66 and 2.10 with these seeds against 1.62 and 2.37 with the main ones.

\paragraph{Ablations.} Each component was removed once, on the exploration set. Completion is what moves the method off the trade-off: against the composite alone it lowers the noise by .41 $[.07,.88]$ and .70 $[.21,1.51]$ on the validation set (a preliminary configuration) without a significant change in response, and on the exploration set its gain concentrates where predicted, in uncovered pixels (speed-up $N$ 14.1$\to$2.0), while the covered-region response is unchanged (.43 vs .42, .34 vs .35). Fusing five sources instead of two lowers the noise by .21 and 1.31. Pasting the covered pixels back matters: without it the noise rises by .33 and .46, because the decoder's rendition of pinned latents loses geometric fidelity along the rollout (covered-region per-pixel error of the planned arm 3.66$\to$4.25, step-wise $\Delta N$ growing from $-.57$ to $+.98$). Treating the arms identically matters as well: shrinking the warp of the planned arms toward re-sampling lowers their own error (4.03$\to$2.95) but raises the paired noise (1.69$\to$2.09), because the logged-speed arm is still warped fully and the two arms' errors no longer cancel in the difference. Placing other agents at their physical time changes the pooled static near-field numbers by at most .13; its effect is pixel-level: in the composite, agent box centres lie at a median .07 and .09 box widths from the physical-time reference, against a criterion of .15. Split by region on the validation set (a preliminary configuration), completion dominates where transport has evidence (87\% of the true change: $\alpha$ .48/.37 for both, $N$ 1.25$\to$1.13 and 1.64$\to$1.39) and trades response for noise where it has none: there the composite has no image, the reader extrapolates depth from the transported surroundings ($\alpha$ 1.06/.69 at $N$ 3.24/5.39), and the completion's own content follows the plan less ($\alpha$ .34/.19 at $N$ 1.14/1.93). Two constraints came from failures: covered pixels must stay the transported ones, and all arms must be processed identically.

\subsection{Speed response without a depth reader}
\label{app:flow}
\paragraph{Construction.} The lateral check (Sec.~\ref{sec:refs}) compares optical flow between two generated arms with the image motion that real geometry dictates, so no learned reader sits on the truth side. We apply the same construction to speed plans. For step $j$ of plan $s$ and the logged-speed arm, each static LiDAR point of the real frame of the rear camera (3--25\,m, outside detected vehicles) is moved with the logged poses to the front camera, which gives its true image displacement $\delta$; the prediction $\hat\delta$ is RAFT~\cite{raft} flow between the two generated frames at the same pixel, and $\alpha$ and $N$ follow Eq.~\ref{eq:direct-response} with the two-dimensional $\hat\delta$ and $\delta$ in place of the depth changes. Two choices were fixed on real video of the exploration contexts only. Points come from the rear camera, because a camera that moves back sees roadside objects the ego has passed, which hide points of the front frame. Only displacements within the range where flow between two real frames reproduces the geometric one are counted, since flow-based scores break down at large frame gaps~\cite{met3r}: up to 40 pixels for all points (real video .98 and 1.00) and 150 for obstacle points (1.01 and 1.00). Contexts without such points at an evaluation step drop out of that step.

\begin{table}[t]
\centering\small\setlength{\tabcolsep}{3pt}
\caption{\textbf{Reader-free speed response} ($\alpha$ / $N$ from image displacement; all five evaluation sets pooled, contexts with points in range).}
\label{tab:flow}
\begin{tabular}{llcc}
\toprule
Points & Arm & $0.5\times$ & $1.5\times$\\
\midrule
all static & real video & .92 / .24 & .96 / .19\\
(87, 89 ctx) & re-sampling & .57 / 5.33 & .54 / 2.98\\
 & ReTime-Compose & .86 / 1.29 & .89 / 1.41\\
\midrule
obstacle & real video & .96 / .05 & .97 / .06\\
(99, 105 ctx) & re-sampling & .70 / .84 & .52 / .38\\
 & ReTime-Compose & .76 / .29 & .74 / .25\\
\bottomrule
\end{tabular}
\end{table}

\paragraph{Controls.} Three checks test whether the flow reader itself could create the effect. Seeds: flow between the two seeds' outputs of the same plan, regressed on the same true displacement, has a slope between $-.11$ and $+.11$ for both arms and speeds, so a change of seed creates no response. Consistency: on points whose forward flow is undone by the backward flow (validation and test sets 1--3; 48--60\% of generated and 74--80\% of real points), ReTime-Compose reads .93/.95, re-sampling .72/.48 and real video .94/.97, and the response gain is $+.21$ $[+.03,+.51]$ and $+.47$ $[+.02,+.63]$. Appearance: blur, noise, JPEG compression and splat holes filled by blur, applied independently to both real frames of exploration contexts, keep the real-video slope between .96 and 1.04, so generator-like artefacts do not bias the ruler; only its noise grows.

\paragraph{Results.} Over all static points the method raises the response over re-sampling by $+.29$ $[-.05,+.83]$ ($P{=}.96$) and $+.35$ $[+.10,+.48]$ and lowers the total error by $4.21$ $[.81,17.9]$ and $1.77$ $[.13,4.92]$ (noise: $4.04$ $[.79,17.2]$ and $1.56$, $P{=}.97$); on obstacle points it lowers the slow-down noise by $.55$ $[.03,2.12]$ and raises the speed-up response by $+.22$ $[+.07,+.33]$ at a total error lower by $.29$ $[.09,.41]$. The sign is robust: leaving out any one drive keeps the response gain between $+.19$ and $+.34$ ($0.5\times$) and between $+.30$ and $+.39$ ($1.5\times$) and the total error lower; the total error is lower in 12 of 15 and 9 of 13 drives, with a median per-context change of $-.18$ and $-.23$, and at the slow-down three contexts carry most of the pooled decrease. Without a depth reader the method's static scene moves .94 and .93 as much as in real video, against .66 and .49 under VGGT: part of the gap to real video under the depth reader is the reader, not the forecast.

\begin{table}[t]
\centering\small\setlength{\tabcolsep}{2.5pt}
\caption{\textbf{Eq.~\ref{eq:rho} under two rulers}: change of the total error $1{-}\mathrm{EF}$ of ReTime-Compose against re-sampling on the same contexts, with 95\% drive-cluster intervals (Epona: all five evaluation sets; GEM and Orbis~2: test sets 3 and 4, whose $\rho$ was computed after their reads). $\rho$: medians of $\Delta$ and $\varepsilon$ over plan steps 15--24 of the same contexts, read with each generator's pose chain (VGGT for Orbis~2). Transport is expected to lose where $\rho$ is well below one.}
\label{tab:rho}
\fitcol{\begin{tabular}{lccc}
\toprule
Generator, plan & $\rho$ & depth reader & no depth reader\\
\midrule
Epona, $0.5\times$ & 1.00 & $-.63$ $[-.93,-.37]$ & $-4.2$ $[-18,-.81]$\\
Epona, $1.5\times$ & 2.67 & $+.02$ $[-.56,+.60]$ & $-1.8$ $[-4.9,-.13]$\\
GEM, $0.5\times$ & .90 & $+.71$ $[-.20,+2.8]$ & $-3.5$ $[-4.0,-1.3]$\\
Orbis~2, $0.5\times$ & .39 & $+.22$ $[-.03,+.79]$ & $+.02$ $[-1.6,+5.8]$\\
Orbis~2, $1.5\times$ & .88 & $+.20$ $[-.11,+.66]$ & $-1.4$ $[-3.8,+5.2]$\\
Epona nuScenes, $0.5\times$ & .34 & noise $+2.2$ $[+.72,+4.0]$ & --\\
\bottomrule
\end{tabular}}
\end{table}

\paragraph{What the ratio compares.} Eq.~\ref{eq:rho} compares the misplacement of re-sampling, $f\Delta/Z$, with the pose part of transport's, $f\varepsilon/Z$. Splatting, depth and completion add to transport's error, while re-sampling carries errors this first-order comparison omits: the nearest rollout frame is also at the wrong time for other agents, and its own content has drifted. Transport therefore loses where $\rho$ is well below one and can win near one or above.

\paragraph{The ratio under both rulers.} Table~\ref{tab:rho} lists the generators and plans of the evaluation sets read with both rulers. Where $\rho$ is near one or above (Epona, 1.00--2.67; GEM, .90) the total error decreases without a depth reader in every case, whereas the depth reader shows no decrease for Epona's speed-up and for GEM; where it is lowest among the cases read with both rulers (Orbis~2's slow-down, .39) neither ruler shows a decrease. The ratio is a data-set-level quantity: per context it does not predict the noise change (Spearman below .2 on the validation set). Where the rulers disagree, the depth reader is unstable on generated content: on the two speed-up contexts that carry most of Epona's excess, its depth for the refined and unrefined versions of the same frames, which differ by 4--15 grey levels on average, differs by a factor of 2--3, after the rollout's content has drifted from the real scene. On nuScenes, 128 further non-overlapping contexts repeat the pattern of Table~\ref{tab:rho}, response up and noise up (response $+.11$ $[+.02,+.20]$, noise $+2.28$ $[+1.14,+3.76]$).

%% file: supp_diag.tex
\section{Trajectory score versus geometric response}
\label{app:metric}
\paragraph{Protocol.} Generated videos, seeds and contexts are those of the KITTI generations of Sec.~\ref{sec:metric}; no model was rerun. The published score is computed by the frozen script of the trajectory read-out (Orbis~2 protocol: VGGT pair poses chained with a LiDAR-anchored scale, least-squares alignment to the command, ADE over 2.4\,s). The response slope $\alpha$, the accuracy $G_1$ and the closeness ratio $G_2$ use our depth reader and the pose-matched LiDAR of each plan on steps where every plan maps to a recorded frame, near field ($\leq$25\,m); $\alpha$ pairs seeds, $G_1$/$G_2$ pool them. Intervals are 2000-sample drive-cluster bootstraps. We use $\alpha$ as the response measure (Sec.~\ref{sec:refs}); $G_1$/$G_2$ are reported as accuracy and closeness descriptors.

\paragraph{Known answer for the published score.} Scoring the real pose-matched video (which executes the command) and the real logged video (which ignores it) against the same command gives scale-aligned ADE ratios ignore/execute of .91 $[.77,1.06]$ at $0.5\times$ and .99 $[.89,1.08]$ at $1.5\times$; without alignment the ratios are 9.37 and 3.83. Scale alignment absorbs a change of speed along a fixed path.

\begin{table*}[t]\centering\footnotesize\setlength{\tabcolsep}{3pt}
\fittext{\begin{tabular}{@{}lccccc@{}}
\toprule
Model & $k_{0.5}$ & $k_{1}$ & $k_{1.5}$ & $\alpha_{0.5}$ & $\alpha_{1.5}$\\
\midrule
DrivingWorld & 1.02 [.82,1.23] & .69 [.59,.76] & .63 [.60,.68] & .17 [.06,.32] & .08 [.02,.15]\\
Vista & .53 [.33,.66] & .69 [.59,.76] & .52 [.47,.60] & $-$.03 [$-$.08,.02] & $-$.05 [$-$.14,.01]\\
Epona & 1.18 [1.15,1.23] & 1.17 [1.12,1.23] & 1.15 [1.07,1.22] & .18 [.06,.37] & .16 [.08,.23]\\
Orbis~2 & 1.42 [1.38,1.47] & 1.10 [1.07,1.14] & .89 [.86,.94] & .23 [.17,.28] & .11 [.09,.14]\\
GEM & .77 [.67,.96] & .83 [.77,.92] & .74 [.72,.83] & .20 [.14,.24] & .10 [.07,.16]\\
\bottomrule
\end{tabular}}
\caption{\textbf{Table~\ref{tab:twolayer} with 95\% drive-cluster intervals.} Kinematic ratio $k_s$ (seed 0, frames 0--24) and geometric slope $\alpha$ (seeds paired); the GEM row uses GEM's own generations (256 contexts, 23 drives).}\label{tab:twolayerci}
\end{table*}

\begin{table*}[t]
\centering\footnotesize\setlength{\tabcolsep}{4pt}
\fittext{\begin{tabular}{@{}lrrrrrrrr@{}}
\toprule
& \multicolumn{4}{c}{$0.5\times$} & \multicolumn{4}{c}{$1.5\times$}\\
Model (contexts) & ADE$_a$ & $\kappa$ & $G_1$ & $G_2$ & ADE$_a$ & $\kappa$ & $G_1$ & $G_2$\\
\midrule
DriveLaW (32) & .82 & .19 & 3.22 & 1.09 & 2.43 & $-$.02 & 3.64 & .81\\
DrivingWorld (59) & .95 & .21 & 3.27 & 1.12 & 1.39 & .42 & 3.63 & .89\\
\quad ft logged / retimed / speed & .44/.54/.45 & $-$.03/.77/.03 & 4.50/3.06/4.36 & 1.02/1.21/.98 & 1.38/1.30/1.33 & .34/.83/.41 & 4.81/4.23/4.54 & .99/1.03/.98\\
Epona (128) & .37 & .96 & 3.90 & 1.14 & 1.32 & .93 & 4.58 & 1.03\\
Orbis 2 (256) & .44 & .64 & 3.34 & 1.14 & .79 & .39 & 4.74 & .96\\
\quad ft logged / retimed (256) & .38/.30 & .32/1.12 & 3.59/3.39 & .97/1.19 & .85/1.08 & .19/.67 & 4.23/5.13 & .88/1.00\\
\quad ftK L0/S0/R1/M1b/M1b$'$ (128) & .48--.54, .35 & .23--1.01 & 2.48--3.32 & .94--1.77 & .63--1.05 & .20--.72 & 3.66--4.66 & .84--1.03\\
Vista (64) & .51 & .71 & 2.89 & 1.33 & 1.23 & .18 & 4.06 & .89\\
\midrule
Real pose-matched video & \multicolumn{4}{c}{$G_2$ 2.1--3.9} & \multicolumn{4}{c}{$G_2$ 1.4--2.4}\\
\bottomrule
\end{tabular}}
\caption{\textbf{Per-model read-out.} ADE$_a$: scale-aligned ADE (m); $\kappa$: travelled-distance response (1 = executes); $G_1$: near-field RMSE (m) to the commanded LiDAR; $G_2=\mathrm{MSE}(\widehat d_s,d_{1.0})/\mathrm{MSE}(\widehat d_s,d_s)$, the closeness ratio ($G_2{=}1$: equidistant from the logged and commanded futures; it does not measure the amount of change). Variant rows list fine-tuning on logged, retimed or speed-augmented pairs; ftK rows are a 128-context screen of fine-tuning recipes. Full intervals will be released with the code.}
\label{tab:metric}
\end{table*}

Table~\ref{tab:directall} lists the direct response of every video model.
\begin{table*}[t]\centering\small\setlength{\tabcolsep}{3pt}
\fittext{\begin{tabular}{@{}lccc@{}}
\toprule
Model (contexts) & $\alpha$, $0.5\times$ & $\alpha$, $1.5\times$ & real video $0.5/1.5$\\
\midrule
DriveLaW$^*$ (32) & +.04 [$-$.01,+.13] & +.06 [$-$.03,+.28] & 0.60 / 0.71\\
DrivingWorld (59) & +.17 [+.06,+.32] & +.08 [+.01,+.15] & 0.65 / 0.71\\
\quad ft logged (59) & +.08 [$-$.02,+.21] & +.17 [+.09,+.26] & 0.65 / 0.71\\
\quad ft retimed (59) & +.28 [+.12,+.50] & +.18 [+.08,+.29] & 0.65 / 0.71\\
\quad ft speed-aug. (59) & +.05 [$-$.06,+.17] & +.05 [$-$.04,+.15] & 0.65 / 0.71\\
Epona (128) & +.18 [+.06,+.38] & +.16 [+.09,+.22] & 0.66 / 0.66\\
Orbis 2 (256) & +.23 [+.17,+.28] & +.11 [+.09,+.14] & 0.73 / 0.70\\
\quad ft logged (256) & +.10 [+.08,+.13] & +.05 [+.02,+.06] & 0.73 / 0.70\\
\quad ft retimed (256) & +.30 [+.24,+.37] & +.09 [+.06,+.15] & 0.73 / 0.70\\
\quad ftK L0 (128) & +.06 [+.01,+.14] & +.05 [+.02,+.07] & 0.66 / 0.66\\
\quad ftK S0 (128) & +.06 [+.04,+.09] & +.02 [+.00,+.05] & 0.66 / 0.66\\
\quad ftK R1 (128) & +.28 [+.21,+.40] & +.14 [+.09,+.19] & 0.66 / 0.66\\
\quad ftK M1b (128) & +.31 [+.22,+.44] & +.16 [+.11,+.22] & 0.66 / 0.66\\
\quad ftK M1b$'$ (128) & +.38 [+.35,+.47] & +.18 [+.15,+.23] & 0.77 / 0.72\\
Vista (64) & $-$.03 [$-$.08,+.02] & $-$.05 [$-$.14,+.01] & 0.64 / 0.71\\
GEM (256) & +.20 [+.14,+.24] & +.10 [+.07,+.16] & 0.73 / 0.70\\
\midrule
Fixed transport (256) & +.54 [+.50,+.61] & +.39 [+.33,+.50] & 0.73 / 0.70\\
\bottomrule
\end{tabular}}
\caption{\textbf{Direct response of every video model.} Paired-change slope $\alpha$ of generated depth against real LiDAR change (Eq.~\ref{eq:direct-response}), seeds paired, near field, 95\% drive-cluster intervals; real-video values use the same reader on the same contexts. $^*$Exploratory adapter (failed a sanity check on the sign of turning). EF is negative for every video model because uncorrelated reader noise in two independently read videos adds to the change error; $\alpha$ is unaffected by this noise in expectation.}\label{tab:directall}
\end{table*}

\paragraph{Accuracy rank statistics.} Over the fifteen models of Table~\ref{tab:metric} (Table~\ref{tab:metric} includes DriveLaW but not GEM; the fifteen of Sec.~\ref{sec:metric} include GEM, not DriveLaW), for the accuracy $G_1$, Kendall $\tau$(ADE$_a$, $G_1$) is $-0.26$ ($p{=}.18$) and $-0.16$ ($p{=}.44$); $\tau$(ADE$_a$, $G_2$) is $+0.05$ and $+0.07$. Unaligned ADE$_m$ gives $\tau$ with $G_2$ of $+0.65$ ($p{=}.001$) and $+0.49$ ($p{=}.011$), and with $G_1$ of $+0.34$ and $-0.50$ ($p{=}.008$). Pairs ranked oppositely by ADE$_a$ and $G_1$ with separated intervals: nine at $0.5\times$ (e.g.\ DrivingWorld logged-tuned versus base, Epona versus retimed Orbis~2 variants) and four at $1.5\times$ (e.g.\ DrivingWorld versus Orbis~2).

\paragraph{Reader independence.} We re-read two variants with an independent single-frame metric reader (Depth-Anything-V2 Metric-VKITTI-Large), fixing one scale per video from frame-0 context LiDAR; criteria were recorded before reading.
\begin{center}\footnotesize\setlength{\tabcolsep}{3pt}
\fitcol{\begin{tabular}{@{}llccc@{}}
\toprule
Model & Reader & $\alpha$, $0.5\times$ & $\alpha$, $1.5\times$ & real video\\
\midrule
DrivingWorld ft logged (59) & DA2 & .12 [.02,.27] & .12 [.06,.24] & .79 / .80\\
 & VGGT & .08 & .17 & .65 / .71\\
Orbis 2 ftK M1b$'$ (128) & DA2 & .53 [.47,.60] & .31 [.22,.36] & .98 / .98\\
 & VGGT & .38 & .18 & .77 / .72\\
\bottomrule
\end{tabular}}
\end{center}
Real-video slopes stay far above zero and the model ordering is unchanged. Relative to real video, the logged-pair DrivingWorld variant stays near .15 of the real slope; the best retimed Orbis~2 variant reaches .49 (VGGT) or .54 (DA2) at $0.5\times$ and .26/.32 at $1.5\times$, so ``about half'' is reader-robust only to this approximation. The closeness ratio $G_2$ agrees in direction under both readers.

\paragraph{A decision quantity.} The failure reaches the quantity a planner comparing speeds needs: how far the nearest obstacle in the ego corridor will be. On contexts with an obstacle within 40\,m whose true clearance differs between the two plans by more than 1\,m (51--177 per model), the forecast clearance change follows the LiDAR one with a slope of .50--.63 for real video but at most .14 for Orbis~2 and DrivingWorld: an evaluator reading the forecast would see at most a seventh of how much closer speeding up brings the obstacle. nuScenes shows the same pattern.

\paragraph{History copying stops at the camera.} We keep the plan at the logged speed and feed the \emph{history} played at $0.5\times$ or $1.5\times$; for static content the correct forecast is unchanged. DrivingWorld's generated camera follows the history about as strongly as a changed plan, its path changing by about a third, the copycat pattern of imitation learning~\cite{dehaan2019causal,copycat}, while the scene geometry follows neither (Table~\ref{tab:copycat}). Epona ignores the retimed history, yet its geometry still under-responds: the general failure is geometric, copying is specific.

\begin{table}[h]
\centering\small\setlength{\tabcolsep}{3.5pt}
\caption{\textbf{DrivingWorld's camera copies a retimed history; its scene follows neither history nor plan.} KITTI, 64 contexts. Path ratio: generated camera-path length over the logged-speed generation (real: .53, 1.47 for a changed plan; 1, 1 for a changed history). Response against the corresponding real future (real video .66--.78).}
\label{tab:copycat}
\fitcol{\begin{tabular}{lcccc}
\toprule
 & \multicolumn{2}{c}{change the plan} & \multicolumn{2}{c}{change the history}\\
 & $0.5\times$ & $1.5\times$ & $0.5\times$ & $1.5\times$\\
\midrule
path ratio & .70--.82 & 1.33--1.35 & .73 & 1.36\\
response $\alpha$ & .15--.26 & .01--.14 & .08 & $-$.08 to $-$.01\\
\bottomrule
\end{tabular}}
\end{table}

\section{Dissecting the noise of the context repair}
\label{app:noise}
All rows use the 32 exploration contexts, one seed, and the measurement of Sec.~\ref{sec:refs}; $N=(1-\mathrm{EF})-(1-\alpha)^2$ for $0.5\times$/$1.5\times$. Each hypothesis and its decision rule were written before the reading.

\begin{table*}[t]
\centering\small\setlength{\tabcolsep}{3pt}
\fittext{\begin{tabular}{llll}
\toprule
Hypothesis & Targeted change & Rule & Reading\\
\midrule
gating without power & trust the gate only where plan/default renders separate & lower $N$ at equal $\alpha$ & $N$ 4.75/5.38\\
OOD context encoding & encode the composite as last frame of a moving clip & $N$ drops $\ge1$ & 4.76/6.34 (gain only)\\
sampling noise & average two seeds & noise drops $\ge35\%$ & 14\%/21\%\\
divergent hole content & holes from the logged-speed plan & --- & holes 0--4\% of tokens\\
non-shared sources & $0.5\times$ plan read from the logged plan at matched pose & --- & $N$ 4.17\\
resampling order & fetch colour from frame 0 first & $N$ drops $\ge1$ & drop .96; local error unchanged\\
\bottomrule
\end{tabular}}
\caption{\textbf{Six explanations, stated in advance, of the context-repair noise}, each with its targeted change, decision rule and reading; none explains half of the gap.}\label{tab:noisehyp}
\end{table*}

\paragraph{Single-component injection.} On real history, injecting one degradation at a time gives $N$ .84/.95 (greedy), .96/.94 (blur), .97/1.33 (shift), 1.02/1.38 (holes) and 2.00/2.39 (last frame replaced by the plan render of frame 0), against .98/.82 without injection and 4.26/4.90 for the repair; no single component explains half of the gap.

\paragraph{Per-plan error decomposition.} With $e$ the predicted minus measured depth of each plan, the paired residual is $e_s-e_1$ and $\sum(e_s-e_1)^2=S_s+S_1-2C$. After dividing each frame's prediction by its median ratio to LiDAR, the local errors $S/\sum\Delta d^2$ ($0.5\times$ plan / logged plan) are .29/.38 for real history, .98/1.25 unmodified, .95/1.17 with the two-hypothesis gate (logged plan untouched), and 1.04--1.06/3.74--4.31 for the ungated repairs; at $1.5\times$ the warped plans reach 3.96--5.75. The reference plan of an ungated repair is itself warped and damaged, which is why gating lowers $N$ only together with $\alpha$.

%% file: supp_forecaster.tex
\section{Fine-tuning the plan channel}
\label{app:retimeresults}
\label{app:retimedetails}
\paragraph{Retimed pairs.} A retimed training example $(H,p_\tau,Y_\tau)$ keeps the context of its logged counterpart and takes the target recorded at the retimed pose; unlike whole-clip speed augmentation, it breaks the correlation between history and plan. In a controlled Orbis~2 study we fine-tune only the plan-receiving conditioning and adaptive-normalization parameters (172\,M parameters, 2000 updates, batch two; retimed speeds $0.4$--$1.6\times$, one sample in four logged). Logged-pair travel is predictable from context kinematics with $R^2{=}.95$, retimed pairs with $.44$, and the one-step camera response rises from .03--.07 under logged, coverage-matched and whole-clip training to .84/.68 under retimed pairs.

\paragraph{What reaches the scene.} Under the direct slope of Sec.~\ref{sec:metric}, the retimed Orbis~2 variant gains at the slow-down (.23 to .30) but not at the speed-up (.11 to .09), and the retimed DrivingWorld variant reaches .28/.18 against .17/.08 for its base (Table~\ref{tab:directall}). Retiming the released model's own logged-speed rollout without any training is as accurate on supported static geometry: on 128 windows from 11 drives, fine-tuning changes the near-field error by $+1.9\%$ $[-3.4,+14.4]$ at $0.5\times$ and $+3.9\%$ $[+2.3,+5.8]$ at $1.5\times$ against linear interpolation of that rollout. Applying the plan to the rollout thus matches what learning it from counterfactual pairs achieves, the observation ReTime-Compose builds on.

\paragraph{Teaching the plan channel to the generator.}\label{app:whynotft} We also distilled paired changes from retimed references into Epona's plan pathway, with a paired loss on a shared noised input and an anchor to the released model. On one context the pathway fits the paired change within 100 steps (first-step paired response .21 to .96), so it can express the response. Across contexts it does not transfer: on 30 held-out validation contexts the first-step paired response, a proxy of the same magnitude as $\alpha$ (released model .25/.17), is .27/.20 after 4000 steps, .27/.19 with a four times larger batch (training contexts .61), and .24/.19 after 5000 steps on 662 training contexts. Per-sample gradients of the paired loss share only 5--7\% of their direction. The plan enters as a global modulation vector while the change it must produce varies per pixel with depth, and the clean history already predicts the logged future; with a few hundred counterfactual contexts the network memorises per-context changes instead, the bias of Sec.~\ref{sec:predictor}.

\section{A geometric forecaster: the plan as a vector or as aligned geometry}
\label{sec:method}
\label{sec:experiments}
Section~\ref{sec:predictor} uses a small depth forecaster trained on KITTI to test where a learned forecaster takes the plan from; the plan is given either as a vector or as geometry aligned with the output.

\paragraph{Model.} A historical point $X_i$ in camera $i$ is transported to the planned camera, $X_p{=}R_{pi}X_i{+}t_{pi}$, $u_p{=}\pi(K_pX_p)$. Historical LiDAR gives sparse target depth $z_p$ with support $m_p$; the frozen context-depth estimate, transported the same way, gives a log-depth prior $\ell_p$ with support $v_p$, and its untransported map is $\ell_c$. A head of about 0.94\,M parameters reads these maps, the context RGB and the measurement--prior offset, and predicts mixture weights $w$ and a residual $r$:
\begin{align}
\widehat d_p&=\exp\{\ell_{\rm ref}+b_p\tanh r\},\label{eq:bound}\\
\ell_{\rm ref}&=m_p\log z_p+(1-m_p)(w_1\ell_c+w_2\ell_p),\nonumber
\end{align}
with bound $b_p=0.1m_p+(1-m_p)[0.3v_p+2.0(1-v_p)]$, so the head applies a bounded correction to the transported evidence. Two target poses share one history, and training minimises the mean of their masked Huber depth losses plus a Huber loss on the paired change between them (the form WorldSync uses for intervention effects~\cite{worldsync}); 6000 AdamW steps, batch eight, learning rate $10^{-3}$ with cosine decay, three seeds.

\paragraph{Where the plan enters.} On five drive-disjoint folds plus a validation fold (31 drives), we regress the forecast depth change on the true change under the plan (plan slope $\alpha$) and on the change implied by the history alone (history slope $\beta$), at $0.5\times$/$1.5\times$:
\begin{center}\small\setlength{\tabcolsep}{4pt}
\begin{tabular}{@{}lcc@{}}
\toprule
Plan given as & history slope $\beta$ & plan slope $\alpha$\\
\midrule
vector, with history & .43 / .41 & .05 / .05\\
vector, learned warp, no history & -- & .53 / .36\\
aligned depth, with history & .06 / .06 & .79 / .79\\
aligned depth, bounded head & -- & .82 / .71\\
\bottomrule
\end{tabular}
\end{center}
Trained on the same retimed pairs, the vector head copies the history and barely uses the plan, while the same plan given as an aligned depth map is used almost fully and the history is ignored: the channel, not the data, decides whether the plan is used.

\paragraph{Accuracy.} On 1187 KITTI windows from 26 drives (five drive-disjoint folds, near field ${\le}25$\,m), the three seeds lower the drive-paired near-field RMSE of fixed transport by 30.9--32.2\%, with every drive improved, and raise the paired-change response (Table~\ref{tab:direct}). On a nuScenes confirmation set frozen before reading (479 contexts from 120 scenes not used for training or development), the KITTI-only checkpoints lower the near-field log-RMSE by 25.2--33.7\% at every plan and seed, each interval upper bound below $-22\%$ (scene-equal paired log-RMSE ratios, the estimand fixed in advance). Against the point-cloud forecaster 4D-Occ~\cite{khurana2023}, near-field ray L1 is .83--.86 of its value at 0.6--1.0\,s on 890 KITTI windows from 23 drives and about 20\% lower at 1\,s on the nuScenes confirmation scenes (1.30--1.33 vs 1.65\,m), with image priors and a 2\,s history (4D-Occ: 0.5\,s, LiDAR only, trained on these nuScenes scenes).

\begin{table}[t]
\centering\small\setlength{\tabcolsep}{3pt}
\begin{tabular}{@{}lrr@{}}
\toprule
Method & $0.6\times$: EF / $\alpha$ & $1.6\times$: EF / $\alpha$\\
\midrule
Fixed transport & $-.04/.46$ & $-.27/.39$\\
Measurement hybrid & $.13/.54$ & $-.10/.46$\\
Source-view completion & $.12/.59$ & $-.14/.52$\\
Target view, no depth prior & $.59/.81$ & $.50/.78$\\
Bounded head (Eq.~\ref{eq:bound}) & $\mathbf{.62/.78}$ & $\mathbf{.55/.74}$\\
\bottomrule
\end{tabular}
\caption{\textbf{Paired-change response of the geometric forecaster.} Near-field common LiDAR support, KITTI five folds, seed 0, relative to the logged plan; pooled EF and slope (Eq.~\ref{eq:direct-response}).}
\label{tab:direct}
\end{table}

\section{Where the forecaster's remaining gap lies}
\label{app:gap}
We split the paired-change records of Table~\ref{tab:direct} by whether the target-view pixel receives a transported LiDAR point; unsupported pixels are 66\% ($0.6\times$) and 84\% ($1.6\times$) of near-field change pixels.
\begin{center}\small\setlength{\tabcolsep}{4pt}
\begin{tabular}{@{}lcc@{}}
\toprule
Slope $\alpha$, supported / unsupported & $0.6\times$ & $1.6\times$\\
\midrule
Fixed transport & .56 / .40 & .54 / .36\\
Execute transported LiDAR only & .70 / .45 & .70 / .40\\
Bounded head & .83 / .75 & .82 / .73\\
Target view, no depth prior & .86 / .78 & .84 / .77\\
Source-view completion & .68 / .54 & .66 / .49\\
\bottomrule
\end{tabular}
\end{center}
The remaining gap is spread over both kinds of pixel in proportion to their share (supported minus unsupported .08/.09). Executing the transported LiDAR exactly reaches only .70 on supported pixels, so part of the attenuation lies in the reference itself (moving content and differences between historical scans and target labels). Learning adds most where no measurement reaches (+.30/+.33 over LiDAR execution, against +.13/+.12 where it does).

\section{From diagnosis to design to ablation}
\label{app:chain}
Each design choice of the context repair (Sec.~\ref{sec:channel}, Appendix~\ref{app:repair}) follows from one measured finding and has one ablation row.

\begin{table*}[t]
\centering\small\setlength{\tabcolsep}{3pt}
\fittext{\begin{tabular}{@{}>{\raggedright\arraybackslash}p{.36\textwidth}>{\raggedright\arraybackslash}p{.28\textwidth}>{\raggedright\arraybackslash}p{.32\textwidth}@{}}
\toprule
Finding & Design & Ablation / reading\\
\midrule
plan tokens contribute nothing once the context is correct ($.61/.59$) & act on the context, not the tokens & token hint $.09/{-}.05$; context warp $.54/.30$ (confirmation)\\
the geometry follows the context frames & warp the context along the plan at every step & first-frame-only evidence $.39/.12$; step-wise $.37/.20$\\
hole content with nearest depth over-magnifies & hole depth from a monocular metric model & $.37/.20\to.54/.26$\\
cost is local error compounding along the chain (3.7--5.8 vs 0.3--1.6) & revise covered tokens softly (soft anchoring) & weight $w{=}8$: .45/.04, $N$ 2.47/1.91; no $w$ serves both speeds\\
adapter trained on clean warps over-trusts the context (global scale shift at $1.5\times$) & train on the model's own rollout contexts (error recycling) & .51/.46 at $N$ 4.99/7.47: on the same trade-off\\
refinement from generated frames loses the plan state & keep the warp chain as the carrier of the plan & refine: $.23/.31$, $.15/.12$, Epona $.22/.10$\\
\bottomrule
\end{tabular}}
\caption{\textbf{From finding to design to ablation} ($\alpha$ at $0.5\times$/$1.5\times$ unless noted).}
\label{tab:chain}
\end{table*}

%% file: main.bbl
\begin{thebibliography}{65}
\providecommand{\natexlab}[1]{#1}
\providecommand{\url}[1]{\texttt{#1}}
\expandafter\ifx\csname urlstyle\endcsname\relax
  \providecommand{\doi}[1]{doi: #1}\else
  \providecommand{\doi}{doi: \begingroup \urlstyle{rm}\Url}\fi

\bibitem[Arai et~al.(2025)Arai, Ishihara, Takahashi, and Yamaguchi]{actbench}
Hidehisa Arai, Keishi Ishihara, Tsubasa Takahashi, and Yu Yamaguchi.
\newblock {{ACT-Bench}: Towards Action Controllable World Models for Autonomous
  Driving}.
\newblock In \emph{ICLR}, 2025.

\bibitem[Asim et~al.(2025)Asim, Wewer, Wimmer, Schiele, and Lenssen]{met3r}
Mohammad Asim, Christopher Wewer, Thomas Wimmer, Bernt Schiele, and Jan~Eric
  Lenssen.
\newblock {{MEt3R}: Measuring Multi-View Consistency in Generated Images}.
\newblock In \emph{CVPR}, 2025.

\bibitem[Bahmani et~al.(2025)Bahmani, Skorokhodov, Qian, Siarohin, Menapace,
  Tagliasacchi, Lindell, and Tulyakov]{ac3d}
Sherwin Bahmani, Ivan Skorokhodov, Guocheng Qian, Aliaksandr Siarohin, Willi
  Menapace, Andrea Tagliasacchi, David~B. Lindell, and Sergey Tulyakov.
\newblock {{AC3D}: Analyzing and Improving 3D Camera Control in Video Diffusion
  Transformers}.
\newblock In \emph{CVPR}, 2025.

\bibitem[Bi{\'n}kowski et~al.(2018)Bi{\'n}kowski, Sutherland, Arbel, and
  Gretton]{kid}
Miko{\l}aj Bi{\'n}kowski, Danica~J. Sutherland, Michael Arbel, and Arthur
  Gretton.
\newblock {Demystifying MMD GANs}.
\newblock In \emph{ICLR}, 2018.

\bibitem[Caesar et~al.(2020)Caesar, Bankiti, Lang, Vora, Liong, Xu, Krishnan,
  Pan, Baldan, and Beijbom]{nuscenes}
Holger Caesar, Varun Bankiti, Alex~H. Lang, Sourabh Vora, Venice~Erin Liong,
  Qiang Xu, Anush Krishnan, Yu Pan, Giancarlo Baldan, and Oscar Beijbom.
\newblock {{nuScenes}: A Multimodal Dataset for Autonomous Driving}.
\newblock In \emph{CVPR}, 2020.

\bibitem[Cai et~al.(2026)Cai, Song, Zhang, et~al.]{whatifworld}
Kunlin Cai, Rui Song, Jinghuai Zhang, et~al.
\newblock {What-If World: A Causal Benchmark for General World Models in
  Embodied Scenarios}.
\newblock \emph{arXiv preprint arXiv:2605.27589}, 2026.

\bibitem[Cao et~al.(2022)Cao, Rockwell, and Johnson]{fwd}
Ang Cao, Chris Rockwell, and Justin Johnson.
\newblock {{FWD}: Real-Time Novel View Synthesis With Forward Warping and
  Depth}.
\newblock In \emph{CVPR}, pages 15713--15724, 2022.

\bibitem[Chen et~al.(2025)Chen, Zheng, Wang, Zhang, Zhan, Jia, Keutzer, and
  Zhang]{geodrive}
Anthony Chen, Wenzhao Zheng, Yida Wang, Xueyang Zhang, Kun Zhan, Peng Jia, Kurt
  Keutzer, and Shanghang Zhang.
\newblock {{GeoDrive}: {3D} Geometry-Informed Driving World Model with Precise
  Action Control}.
\newblock \emph{arXiv preprint arXiv:2505.22421}, 2025.

\bibitem[Chen et~al.(2026)Chen, Liu, Wu, Guo, Wang, Qian, Chen, Cao, Tang, and
  Zhang]{worldsync}
Sixiang Chen, Jiaming Liu, Jixian Wu, Yichen Guo, Tinghao Wang, Siyuan Qian,
  Hao Chen, Jiajun Cao, Jian Tang, and Shanghang Zhang.
\newblock {Do Robotic World Models Really Follow Actions? Diagnosing and
  Aligning Action-Conditioned Generation for Policy Learning}.
\newblock \emph{arXiv preprint arXiv:2608.24885}, 2026.

\bibitem[de~Haan et~al.(2019)de~Haan, Jayaraman, and Levine]{dehaan2019causal}
Pim de Haan, Dinesh Jayaraman, and Sergey Levine.
\newblock {Causal Confusion in Imitation Learning}.
\newblock In \emph{NeurIPS}, 2019.

\bibitem[Ding et~al.(2026)Ding, Chen, Cai, Xu, Wang, Lu, Li, Chen, Gao, Tao,
  Wan, and Zhao]{playworld}
Kaixin Ding, Xi Chen, Minghong Cai, Zhiyuan Xu, Yiyang Wang, Yuxiang Lu, Junyi
  Li, Shuyang Chen, Yuan Gao, Xin Tao, Pengfei Wan, and Hengshuang Zhao.
\newblock {{PlayWorld}: Benchmarking World Models with Agent Players over
  Long-Horizon Objectives}.
\newblock \emph{arXiv preprint arXiv:2608.13552}, 2026.

\bibitem[Gao et~al.(2024)Gao, Yang, Chen, Chitta, Qiu, Geiger, Zhang, and
  Li]{vista}
Shenyuan Gao, Jiazhi Yang, Li Chen, Kashyap Chitta, Yihang Qiu, Andreas Geiger,
  Jun Zhang, and Hongyang Li.
\newblock {Vista: A Generalizable Driving World Model with High Fidelity and
  Versatile Controllability}.
\newblock In \emph{NeurIPS}, 2024.

\bibitem[Geiger et~al.(2013)Geiger, Lenz, Stiller, and Urtasun]{kitti}
Andreas Geiger, Philip Lenz, Christoph Stiller, and Raquel Urtasun.
\newblock {Vision meets Robotics: The {KITTI} Dataset}.
\newblock \emph{International Journal of Robotics Research}, 32\penalty0
  (11):\penalty0 1231--1237, 2013.

\bibitem[Hassan et~al.(2025)Hassan, Stapf, Rahimi, Rezende, Haghighi,
  Br{\"u}ggemann, Katircioglu, Zhang, Chen, Saha, Cannici, Aljalbout, Ye, Wang,
  Davtyan, Salzmann, Scaramuzza, Pollefeys, Favaro, and Alahi]{gem}
Mariam Hassan, Sebastian Stapf, Ahmad Rahimi, Pedro M.~B. Rezende, Yasaman
  Haghighi, David Br{\"u}ggemann, Isinsu Katircioglu, Lin Zhang, Xiaoran Chen,
  Suman Saha, Marco Cannici, Elie Aljalbout, Botao Ye, Xi Wang, Aram Davtyan,
  Mathieu Salzmann, Davide Scaramuzza, Marc Pollefeys, Paolo Favaro, and
  Alexandre Alahi.
\newblock {{GEM}: A Generalizable Ego-Vision Multimodal World Model for
  Fine-Grained Ego-Motion, Object Dynamics, and Scene Composition Control}.
\newblock In \emph{CVPR}, 2025.

\bibitem[He et~al.(2025)He, Xu, Guo, Wetzstein, Dai, Li, and Yang]{cameractrl}
Hao He, Yinghao Xu, Yuwei Guo, Gordon Wetzstein, Bo Dai, Hongsheng Li, and
  Ceyuan Yang.
\newblock {{CameraCtrl}: Enabling Camera Control for Text-to-Video Generation}.
\newblock In \emph{ICLR}, 2025.

\bibitem[Hu et~al.(2024)Hu, Yin, Jia, Deng, Guo, Zhang, Long, and
  Tan]{drivingworld}
Xiaotao Hu, Wei Yin, Mingkai Jia, Junyuan Deng, Xiaoyang Guo, Qian Zhang,
  Xiaoxiao Long, and Ping Tan.
\newblock {DrivingWorld: Constructing World Model for Autonomous Driving via
  Video GPT}.
\newblock \emph{arXiv preprint arXiv:2412.19505}, 2024.

\bibitem[Huang et~al.(2025)Huang, Li, He, Zhou, and Shechtman]{selfforcing}
Xun Huang, Zhengqi Li, Guande He, Mingyuan Zhou, and Eli Shechtman.
\newblock {Self Forcing: Bridging the Train-Test Gap in Autoregressive Video
  Diffusion}.
\newblock \emph{arXiv preprint arXiv:2506.08009}, 2025.

\bibitem[Jayasumana et~al.(2024)Jayasumana, Ramalingam, Veit, Glasner,
  Chakrabarti, and Kumar]{rethinkfid}
Sadeep Jayasumana, Srikumar Ramalingam, Andreas Veit, Daniel Glasner, Ayan
  Chakrabarti, and Sanjiv Kumar.
\newblock {Rethinking {FID}: Towards a Better Evaluation Metric for Image
  Generation}.
\newblock In \emph{CVPR}, 2024.

\bibitem[Khanzada et~al.(2026)Khanzada, Sridhar, and Kwon]{branchdrive}
Feeza~Khan Khanzada, Sudarshan Sridhar, and Jaerock Kwon.
\newblock {{BranchDrive}: A Branch-Structured Dataset for Action-Conditioned
  Driving Prediction}.
\newblock \emph{arXiv preprint arXiv:2609.27275}, 2026.

\bibitem[Khurana et~al.(2023)Khurana, Hu, Held, and Ramanan]{khurana2023}
Tarasha Khurana, Peiyun Hu, David Held, and Deva Ramanan.
\newblock {Point Cloud Forecasting as a Proxy for {4D} Occupancy Forecasting}.
\newblock In \emph{CVPR}, 2023.

\bibitem[Kong et~al.(2025)Kong, Yang, Zheng, and Wang]{worldwarp}
Hanyang Kong, Xingyi Yang, Xiaoxu Zheng, and Xinchao Wang.
\newblock {{WorldWarp}: Propagating 3D Geometry with Asynchronous Video
  Diffusion}.
\newblock \emph{arXiv preprint arXiv:2512.19678}, 2025.

\bibitem[Li et~al.(2026)Li, Pan, Luan, Gao, and Alahi]{svi}
Wuyang Li, Wentao Pan, Po-Chien Luan, Yang Gao, and Alexandre Alahi.
\newblock {Stable Video Infinity: Infinite-Length Video Generation with Error
  Recycling}.
\newblock In \emph{ICLR}, 2026.

\bibitem[Liang et~al.(2026)Liang, Kong, Yan, Liu, Yang, Huang, Yin, Zuo, Hu,
  Zhu, Lu, Liu, Jiang, Li, Li, Zhuo, Ng, Cottereau, Gao, Pan, Ooi, and
  Liu]{worldlens}
Ao Liang, Lingdong Kong, Tianyi Yan, Hongsi Liu, Wesley Yang, Ziqi Huang, Wei
  Yin, Jialong Zuo, Yixuan Hu, Dekai Zhu, Dongyue Lu, Youquan Liu, Guangfeng
  Jiang, Linfeng Li, Xiangtai Li, Long Zhuo, Lai~Xing Ng, Benoit~R. Cottereau,
  Changxin Gao, Liang Pan, Wei~Tsang Ooi, and Ziwei Liu.
\newblock {{WorldLens}: Full-Spectrum Evaluations of Driving World Models in
  Real World}.
\newblock In \emph{CVPR}, 2026.

\bibitem[Lu et~al.(2026)Lu, Wang, Wang, et~al.]{helloworld}
Fan Lu, Hanshi Wang, Zijing Wang, et~al.
\newblock {{HelloWorld}: Towards Practical Applications of Generative Driving
  World Models}.
\newblock \emph{arXiv preprint arXiv:2609.28931}, 2026.

\bibitem[Lugmayr et~al.(2022)Lugmayr, Danelljan, Romero, Yu, Timofte, and
  Van~Gool]{repaint}
Andreas Lugmayr, Martin Danelljan, Andres Romero, Fisher Yu, Radu Timofte, and
  Luc Van~Gool.
\newblock {{RePaint}: Inpainting Using Denoising Diffusion Probabilistic
  Models}.
\newblock In \emph{CVPR}, 2022.

\bibitem[Mittal et~al.(2026)Mittal, Mousakhan, Galesso, Farid, Dienert, Sahay,
  and Brox]{orbis2}
Sudhanshu Mittal, Arian Mousakhan, Silvio Galesso, Karim Farid, Johannes
  Dienert, Rajat Sahay, and Thomas Brox.
\newblock {Orbis 2: A Hierarchical World Model for Driving}.
\newblock \emph{arXiv preprint arXiv:2607.15898}, 2026.

\bibitem[Mousakhan et~al.(2025)Mousakhan, Mittal, Galesso, Farid, and
  Brox]{orbis}
Arian Mousakhan, Sudhanshu Mittal, Silvio Galesso, Karim Farid, and Thomas
  Brox.
\newblock {Orbis: Overcoming Challenges of Long-Horizon Prediction in Driving
  World Models}.
\newblock In \emph{NeurIPS}, 2025.

\bibitem[Ni et~al.(2025{\natexlab{a}})Ni, Zhao, Wang, Zhu, Qin, Huang, Liu,
  Chen, Wang, Zhang, Zhan, Zhan, Jia, Lang, Wang, and Mei]{recondreamer}
Chaojun Ni, Guosheng Zhao, Xiaofeng Wang, Zheng Zhu, Wenkang Qin, Guan Huang,
  Chen Liu, Yuyin Chen, Yida Wang, Xueyang Zhang, Yifei Zhan, Kun Zhan, Peng
  Jia, Xianpeng Lang, Xingang Wang, and Wenjun Mei.
\newblock {{ReconDreamer}: Crafting World Models for Driving Scene
  Reconstruction via Online Restoration}.
\newblock In \emph{CVPR}, 2025{\natexlab{a}}.

\bibitem[Ni et~al.(2025{\natexlab{b}})Ni, Du, Hou, Wu, and Yang]{paralane}
Ziqian Ni, Sicong Du, Zhenghua Hou, Chenming Wu, and Sheng Yang.
\newblock {{Para-Lane}: Multi-Lane Dataset Registering Parallel Scans for
  Benchmarking Novel View Synthesis}.
\newblock \emph{arXiv preprint arXiv:2502.15635}, 2025{\natexlab{b}}.

\bibitem[Oquab et~al.(2024)Oquab, Darcet, Moutakanni, et~al.]{dinov2}
Maxime Oquab, Timoth{\'e}e Darcet, Th{\'e}o Moutakanni, et~al.
\newblock {{DINOv2}: Learning Robust Visual Features without Supervision}.
\newblock \emph{Transactions on Machine Learning Research}, 2024.

\bibitem[Ren et~al.(2025)Ren, Shen, Huang, Ling, Lu, Nimier-David, M{\"u}ller,
  Keller, Fidler, and Gao]{gen3c}
Xuanchi Ren, Tianchang Shen, Jiahui Huang, Huan Ling, Yifan Lu, Merlin
  Nimier-David, Thomas M{\"u}ller, Alexander Keller, Sanja Fidler, and Jun Gao.
\newblock {{GEN3C}: {3D}-Informed World-Consistent Video Generation with
  Precise Camera Control}.
\newblock In \emph{CVPR}, 2025.

\bibitem[Schwarzkopf et~al.(2026)Schwarzkopf, Immel, Blumberg,
  et~al.]{kitscenes}
Richard Schwarzkopf, Fabian Immel, Alexander Blumberg, et~al.
\newblock {The Road Ahead in Autonomous Driving: The {KITScenes} Multimodal
  Dataset}.
\newblock \emph{arXiv preprint arXiv:2606.02956}, 2026.

\bibitem[Shah et~al.(2020)Shah, Tamuly, Raghunathan, Jain, and
  Netrapalli]{shah2020pitfalls}
Harshay Shah, Kaustav Tamuly, Aditi Raghunathan, Prateek Jain, and Praneeth
  Netrapalli.
\newblock {The Pitfalls of Simplicity Bias in Neural Networks}.
\newblock In \emph{NeurIPS}, 2020.

\bibitem[Shi et~al.(2026)Shi, Chu, Wei, Hao, Liu, and Fu]{coco2026}
Yuhong Shi, Zhenhao Chu, Jie Wei, Jun Hao, Jianyi Liu, and Jingwen Fu.
\newblock {Overcoming Statistical Bias in Action-Controllable World Models}.
\newblock \emph{arXiv preprint arXiv:2608.04653}, 2026.

\bibitem[Singh et~al.(2026)Singh, Badola, Kumari, Nagahara, and
  Raman]{camtrolpp}
Prajwal Singh, Arjun Badola, Seema Kumari, Hajime Nagahara, and Shanmuganathan
  Raman.
\newblock {Stabilizing Camera-Controlled Novel View Synthesis at Inference
  Time}.
\newblock \emph{arXiv preprint arXiv:2609.03639}, 2026.

\bibitem[Song et~al.(2025)Song, Yang, Zhao, Li, and Zhang]{worldforge}
Chenxi Song, Yanming Yang, Tong Zhao, Ruibo Li, and Chi Zhang.
\newblock {{WorldForge}: Unlocking Emergent {3D/4D} Generation in Video
  Diffusion Model via Training-Free Guidance}.
\newblock \emph{arXiv preprint arXiv:2509.15130}, 2025.

\bibitem[Song et~al.(2026)Song, Yang, and Zhang]{worldinworld}
Chenxi Song, Yanming Yang, and Chi Zhang.
\newblock {World in World: Explore the World with World Models}.
\newblock \emph{arXiv preprint arXiv:2609.11548}, 2026.

\bibitem[Stein et~al.(2023)Stein, Cresswell, Hosseinzadeh, Sui, Ross,
  Villecroze, Liu, Caterini, Taylor, and Loaiza-Ganem]{stein2023exposing}
George Stein, Jesse~C. Cresswell, Rasa Hosseinzadeh, Yi Sui, Brendan~Leigh
  Ross, Valentin Villecroze, Zhaoyan Liu, Anthony~L. Caterini, J.~Eric~T.
  Taylor, and Gabriel Loaiza-Ganem.
\newblock {Exposing Flaws of Generative Model Evaluation Metrics and Their
  Unfair Treatment of Diffusion Models}.
\newblock In \emph{NeurIPS}, 2023.

\bibitem[Su et~al.(2026)Su, Diermeyer, and Lienkamp]{norminject}
Xiyan Su, Frank Diermeyer, and Markus Lienkamp.
\newblock {Is Energy Guidance All You Need? Training-Free Norm Injection for
  Driving World Models}.
\newblock \emph{arXiv preprint arXiv:2607.10781}, 2026.

\bibitem[Teed and Deng(2020)]{raft}
Zachary Teed and Jia Deng.
\newblock {{RAFT}: Recurrent All-Pairs Field Transforms for Optical Flow}.
\newblock In \emph{ECCV}, 2020.

\bibitem[Tucker and Snavely(2020)]{tucker2020}
Richard Tucker and Noah Snavely.
\newblock {Single-View View Synthesis with Multiplane Images}.
\newblock In \emph{CVPR}, 2020.

\bibitem[Uhrig et~al.(2017)Uhrig, Schneider, Schneider, Franke, Brox, and
  Geiger]{kittidepth}
Jonas Uhrig, Nick Schneider, Lukas Schneider, Uwe Franke, Thomas Brox, and
  Andreas Geiger.
\newblock {Sparsity Invariant {CNNs}}.
\newblock In \emph{3DV}, 2017.

\bibitem[Valevski et~al.(2025)Valevski, Leviathan, Arar, and
  Fruchter]{gamengen}
Dani Valevski, Yaniv Leviathan, Moab Arar, and Shlomi Fruchter.
\newblock {Diffusion Models Are Real-Time Game Engines}.
\newblock In \emph{ICLR}, 2025.

\bibitem[Wang et~al.(2025{\natexlab{a}})Wang, Chen, Karaev, Vedaldi, Rupprecht,
  and Novotny]{vggt}
Jianyuan Wang, Minghao Chen, Nikita Karaev, Andrea Vedaldi, Christian
  Rupprecht, and David Novotny.
\newblock {{VGGT}: Visual Geometry Grounded Transformer}.
\newblock In \emph{CVPR}, 2025{\natexlab{a}}.

\bibitem[Wang et~al.(2025{\natexlab{b}})Wang, Fan, Wang, Chen, and
  Zhang]{freevs}
Qitai Wang, Lue Fan, Yuqi Wang, Yuntao Chen, and Zhaoxiang Zhang.
\newblock {{FreeVS}: Generative View Synthesis on Free Driving Trajectory}.
\newblock In \emph{ICLR}, 2025{\natexlab{b}}.

\bibitem[Wang et~al.(2025{\natexlab{c}})Wang, Xu, Dai, Xiang, Deng, Tong, and
  Yang]{moge}
Ruicheng Wang, Sicheng Xu, Cassie Dai, Jianfeng Xiang, Yu Deng, Xin Tong, and
  Jiaolong Yang.
\newblock {MoGe: Unlocking Accurate Monocular Geometry Estimation for
  Open-Domain Images with Optimal Training Supervision}.
\newblock In \emph{CVPR}, pages 5261--5271, 2025{\natexlab{c}}.

\bibitem[Wang et~al.(2026{\natexlab{a}})Wang, Du, Chen, et~al.]{vehdyn}
Tianyi Wang, Wangsheng Du, Jiazhou Chen, et~al.
\newblock {{VehDyn}: A Driving World Model Benchmark for Vehicle Dynamics}.
\newblock \emph{arXiv preprint arXiv:2609.33264}, 2026{\natexlab{a}}.

\bibitem[Wang and He(2026)]{warpashistory}
Yifan Wang and Tong He.
\newblock {Warp-as-History: Generalizable Camera-Controlled Video Generation
  from One Training Video}.
\newblock \emph{arXiv preprint arXiv:2605.15182}, 2026.

\bibitem[Wang et~al.(2026{\natexlab{b}})Wang, Ye, Su, Xu, and Zhu]{gibbslens}
Yuzhu Wang, Xi Ye, Duo Su, Yangyang Xu, and Jun Zhu.
\newblock {{GibbsLens}: Training-Free Camera Control via Block-Conditional
  {Gibbs} Refinement}.
\newblock In \emph{NeurIPS}, 2026{\natexlab{b}}.

\bibitem[Wang et~al.(2025{\natexlab{d}})Wang, Xia, Bie, Liu, Yu, Bian, and
  Wang]{georeward}
Zhaoqing Wang, Xiaobo Xia, Zhuolin Bie, Jinlin Liu, Dongdong Yu, Jia-Wang Bian,
  and Changhu Wang.
\newblock {Taming Camera-Controlled Video Generation with Verifiable Geometry
  Reward}.
\newblock \emph{arXiv preprint arXiv:2512.02870}, 2025{\natexlab{d}}.

\bibitem[Wei et~al.(2026)Wei, Zhou, Mao, et~al.]{cdlam}
Yufan Wei, Kun Zhou, Lingjun Mao, et~al.
\newblock {Causally Debiased Latent Action Model for Embodied
  Action-Conditioned World Models}.
\newblock \emph{arXiv preprint arXiv:2607.09185}, 2026.

\bibitem[Wen et~al.(2020)Wen, Lin, Darrell, Jayaraman, and Gao]{copycat}
Chuan Wen, Jierui Lin, Trevor Darrell, Dinesh Jayaraman, and Yang Gao.
\newblock {Fighting Copycat Agents in Behavioral Cloning from Observation
  Histories}.
\newblock In \emph{NeurIPS}, 2020.

\bibitem[Wilson et~al.(2021)Wilson, Qi, Agarwal, Lambert, Singh, Khandelwal,
  Pan, Kumar, Hartnett, Pontes, Ramanan, Carr, and Hays]{av2}
Benjamin Wilson, William Qi, Tanmay Agarwal, John Lambert, Jagjeet Singh,
  Siddhesh Khandelwal, Bowen Pan, Ratnesh Kumar, Andrew Hartnett,
  Jhony~Kaesemodel Pontes, Deva Ramanan, Peter Carr, and James Hays.
\newblock {{Argoverse 2}: Next Generation Datasets for Self-Driving Perception
  and Forecasting}.
\newblock In \emph{NeurIPS Datasets and Benchmarks}, 2021.

\bibitem[Wu et~al.(2025)Wu, Zhang, Turki, Ren, Gao, Shou, Fidler, Gojcic, and
  Ling]{difix3d}
Jay~Zhangjie Wu, Yuxuan Zhang, Haithem Turki, Xuanchi Ren, Jun Gao, Mike~Zheng
  Shou, Sanja Fidler, Zan Gojcic, and Huan Ling.
\newblock {{Difix3D+}: Improving {3D} Reconstructions with Single-Step
  Diffusion Models}.
\newblock In \emph{CVPR}, 2025.

\bibitem[Wu et~al.(2026)Wu, Luo, Zhu, et~al.]{fastslow}
Yen-Siang Wu, Rundong Luo, Jingsen Zhu, et~al.
\newblock {Seeing Fast and Slow: Learning the Flow of Time in Videos}.
\newblock \emph{arXiv preprint arXiv:2604.21931}, 2026.

\bibitem[Xi et~al.(2026)Xi, Guo, Hou, Liu, Fang, Liu, and Yang]{geov2v}
Yuchen Xi, Yuhan Guo, Chenwei Hou, Zixu Liu, Jin Fang, Jason Liu, and Ruigang
  Yang.
\newblock {{GeoV2V}: Geometry-Grounded Video Diffusion Model for Driving Scene
  Generation}.
\newblock \emph{arXiv preprint arXiv:2610.25245}, 2026.

\bibitem[Xia et~al.(2026)Xia, Li, Zhou, Yao, Xiong, Sun, Wang, Ma, Chen, Ye,
  Liu, and Wang]{drivelaw}
Tianze Xia, Yongkang Li, Lijun Zhou, Jingfeng Yao, Kaixin Xiong, Haiyang Sun,
  Bing Wang, Kun Ma, Guang Chen, Hangjun Ye, Wenyu Liu, and Xinggang Wang.
\newblock {DriveLaW: Unifying Planning and Video Generation in a Latent Driving
  World}.
\newblock In \emph{CVPR}, 2026.

\bibitem[Xu et~al.(2026)Xu, Zhong, Shu, Jia, Li, Bian, Zhang, Zhang, Xie, Yang,
  and Yin]{eponav2}
Jiawei Xu, Zhizhou Zhong, Zhijian Shu, Mingkai Jia, Mingxiao Li, Jia-Wang Bian,
  Qian Zhang, Kaicheng Zhang, Jin Xie, Jian Yang, and Wei Yin.
\newblock {{EponaV2}: Driving World Model with Comprehensive Future Reasoning}.
\newblock \emph{arXiv preprint arXiv:2605.14696}, 2026.

\bibitem[Yan et~al.(2025)Yan, Xu, Lin, Jin, Guo, Wang, Zhan, Lang, Bao, Zhou,
  and Peng]{streetcrafter}
Yunzhi Yan, Zhen Xu, Haotong Lin, Haian Jin, Haoyu Guo, Yida Wang, Kun Zhan,
  Xianpeng Lang, Hujun Bao, Xiaowei Zhou, and Sida Peng.
\newblock {{StreetCrafter}: Street View Synthesis with Controllable Video
  Diffusion Models}.
\newblock In \emph{CVPR}, 2025.

\bibitem[Yang et~al.(2024)Yang, Kang, Huang, Zhao, Xu, Feng, and Zhao]{da2}
Lihe Yang, Bingyi Kang, Zilong Huang, Zhen Zhao, Xiaogang Xu, Jiashi Feng, and
  Hengshuang Zhao.
\newblock {Depth Anything {V2}}.
\newblock In \emph{NeurIPS}, 2024.

\bibitem[You et~al.(2025)You, Zhu, Liu, and Hou]{nvssolver}
Meng You, Zhiyu Zhu, Hui Liu, and Junhui Hou.
\newblock {{NVS-Solver}: Video Diffusion Model as Zero-Shot Novel View
  Synthesizer}.
\newblock In \emph{ICLR}, 2025.

\bibitem[Zhang et~al.(2026)Zhang, Cui, Xu, Ye, Zhang, and Wang]{cfwm2026}
Jiaru Zhang, Can Cui, Yi Xu, Xin Ye, Ruqi Zhang, and Ziran Wang.
\newblock {How Can Driving World Models Do Counterfactual Prediction?}
\newblock \emph{arXiv preprint arXiv:2608.11601}, 2026.

\bibitem[Zhang et~al.(2025)Zhang, Tang, Hu, Pan, Guo, Liu, Huang, Yuan, Zhang,
  Long, Cao, and Yin]{epona}
Kaiwen Zhang, Zhenyu Tang, Xiaotao Hu, Xingang Pan, Xiaoyang Guo, Yuan Liu,
  Jingwei Huang, Li Yuan, Qian Zhang, Xiao-Xiao Long, Xun Cao, and Wei Yin.
\newblock {{Epona}: Autoregressive Diffusion World Model for Autonomous
  Driving}.
\newblock In \emph{ICCV}, pages 27220--27230, 2025.

\bibitem[Zhao et~al.(2025)Zhao, Ni, Wang, Zhu, Zhang, Wang, Huang, Chen, Wang,
  Zhang, Mei, and Wang]{drivedreamer4d}
Guosheng Zhao, Chaojun Ni, Xiaofeng Wang, Zheng Zhu, Xueyang Zhang, Yida Wang,
  Guan Huang, Xinze Chen, Boyuan Wang, Youyi Zhang, Wenjun Mei, and Xingang
  Wang.
\newblock {{DriveDreamer4D}: World Models Are Effective Data Machines for {4D}
  Driving Scene Representation}.
\newblock In \emph{CVPR}, 2025.

\bibitem[Zhou et~al.(2026)Zhou, Shao, Wang, Zong, Li, and
  Waslander]{drivinggen}
Yang Zhou, Hao Shao, Letian Wang, Zhuofan Zong, Hongsheng Li, and Steven~L.
  Waslander.
\newblock {{DrivingGen}: A Comprehensive Benchmark for Generative Video World
  Models in Autonomous Driving}.
\newblock In \emph{ICLR}, 2026.

\end{thebibliography}
